%% file: main.tex
\documentclass[sigconf]{acmart}
\acmDOI{} % Remove DOI
\acmISBN{} % Remove ISBN
\setcopyright{none} % Remove copyright

\AtBeginDocument{%
  }

\acmYear{Published version's DOI: \url{https://doi.org/10.1145/3580305.3599378}}
\acmConference[Published in ACM KDD '23]{Proceedings of the 29th ACM SIGKDD Conference on Knowledge Discovery and Data Mining}{August 6--10, 2023}{Long Beach, CA, USA.}
\usepackage[labelformat=simple]{subcaption}
\usepackage{amsmath}
\usepackage{amsfonts}
\usepackage{bm}
\usepackage{booktabs}
\usepackage{xcolor}
\usepackage{multirow}
\usepackage{makecell}
\usepackage{amsthm}
\usepackage[inline,shortlabels]{enumitem}

\usepackage{balance}

\DeclareMathOperator*{\argmin}{argmin}   % Jan Hlavacek
\DeclareMathOperator*{\mean}{mean}   % Jan Hlavacek

\begin{document}

% Title

% Your title must be in mixed case, not sentence case.
% That means all verbs (including short verbs like be, is, using,and go),
% nouns, adverbs, adjectives should be capitalized, including both words in hyphenated terms, while
% articles, conjunctions, and prepositions are lower case unless they
% directly follow a colon or long dash
\title{Hierarchical Proxy Modeling for Improved HPO\\ in Time Series Forecasting}

%%
%% The "author" command and its associated commands are used to define
%% the authors and their affiliations.
%% Of note is the shared affiliation of the first two authors, and the
%% "authornote" and "authornotemark" commands
%% used to denote shared contribution to the research.
\author{Arindam Jati}
% \authornote{Both authors contributed equally to this research.}
\email{arindam.jati@ibm.com}
\orcid{0000-0002-9498-8536}
% \authornotemark[1]
\affiliation{%
  \institution{IBM Research}
  \city{Bangalore}
  \country{India}
}

\author{Vijay Ekambaram}
\email{vijaye12@in.ibm.com}
\orcid{0009-0004-5824-9826}
% \authornotemark[1]
\affiliation{
  \institution{IBM Research}
  \city{Bangalore}
  \country{India}
}

\author{Shaonli Pal}
\authornote{The author was in IBM Research while the work was done.}
\email{palshaonli1234@gmail.com}
\orcid{0009-0005-5893-3908}
% \authornotemark[1]
\affiliation{
  \institution{Indian Institute of Technology}
  \city{Jodhpur}
  \country{India}
}

\author{Brian Quanz}
\email{blquanz@us.ibm.com}
\orcid{0000-0002-4136-5538}
% \authornotemark[1]
\affiliation{
  \institution{IBM Research}
  \city{Yorktown Heights, NY}
  \country{USA}
}

\author{Wesley M. Gifford}
\email{wmgifford@us.ibm.com}
\orcid{0000-0003-3678-8410}
% \authornotemark[1]
\affiliation{
  \institution{IBM Research}
  \city{Yorktown Heights, NY}
  \country{USA}
}

\author{Pavithra Harsha}
\email{pharsha@us.ibm.com}
\orcid{0000-0002-6049-7739}
% \authornotemark[1]
\affiliation{
  \institution{IBM Research}
  \city{Yorktown Heights, NY}
  \country{USA}
}

\author{Stuart Siegel}
\email{stus@us.ibm.com}
\orcid{0009-0006-7960-9186}
% \authornotemark[1]
\affiliation{
  \institution{IBM Research}
  \city{Yorktown Heights, NY}
  \country{USA}
}

\author{Sumanta Mukherjee}
\email{sumanm03@in.ibm.com}
\orcid{0000-0003-2042-9966}
% \authornotemark[1]
\affiliation{
  \institution{IBM Research}
  \city{Bangalore}
  \country{India}
}

\author{Chandra Narayanaswami}
\email{chandras@us.ibm.com}
\orcid{0000-0002-4242-5352}
% \authornotemark[1]
\affiliation{
  \institution{IBM Research}
  \city{Yorktown Heights, NY}
  \country{USA}
}

%%
%% By default, the full list of authors will be used in the page
%% headers. Often, this list is too long, and will overlap
%% other information printed in the page headers. This command allows
%% the author to define a more concise list
%% of authors' names for this purpose.
% \renewcommand{\shortauthors}{Jati et al.}
\renewcommand{\shortauthors}{Arindam Jati et al.}

% New commands

%Arindam
\newcommand{\ie}{i.e., }
\newcommand{\eg}{e.g., }
\newcommand{\etc}{\textit{etc.}}
\newcommand{\viz}{\textit{viz. }}
\newcommand{\etal}{\textit{et. al.}}
\newcommand{\hpro}{\texttt{H-Pro}}
% datasets shorcuts
\newcommand{\tour}{\texttt{Tourism}}
\newcommand{\tourL}{\texttt{Tourism-L}}
\newcommand{\wiki}{\texttt{Wiki}}
\newcommand{\mfive}{\texttt{M5}}
\newcommand{\traffic}{\texttt{Traffic}}

% models shorcuts
\newcommand{\deepar}{\texttt{DeepAR}}
\newcommand{\lgbm}{\texttt{LightGBM}}
\newcommand{\thetamodel}{\texttt{Theta}}
\newcommand{\nbeats}{\texttt{N-BEATS}}
\newcommand{\arima}{\texttt{ARIMA}}
\newcommand{\ets}{\texttt{ETS}}
\newcommand{\mint}{\texttt{MinT}}
\newcommand{\mintshr}{\texttt{MinT-shr}}
\newcommand{\mintols}{\texttt{MinT-ols}}
\newcommand{\erm}{\texttt{ERM}}
\newcommand{\permbu}{\texttt{PERMBU}}
\newcommand{\bu}{\texttt{BU}}
\newcommand{\naivebu}{\texttt{BU}}
\newcommand{\td}{\texttt{TD}}
\newcommand{\mo}{\texttt{MO}}
\newcommand{\deepvarplus}{\texttt{DeepVAR+}}
\newcommand{\hieretoe}{\texttt{HierE2E}}
\newcommand{\tcv}{\texttt{TCV}}
\newcommand{\lowest}{\texttt{Lowest}}
\newcommand{\hier}{\texttt{Hier}}
\newcommand{\po}{\texttt{PO}}
\newcommand{\toplevel}{\texttt{Top}}
\newcommand{\avg}{\texttt{Avg}}

% Arindam equations shorthands
\newcommand{\dset}[2]{\mathcal{X}^{#1}_{\mathrm{#2}}}
\newcommand{\algo}[1]{\mathcal{A}\left( \mathcal{X}^L_{\mathrm{#1}};\lambda \right)}
\newcommand{\xlj}{x_t^{l,j}}
\newcommand{\xhatlj}{\hat{x}_t^{l,j}}
\newcommand{\xtildelj}{\Tilde{x}_t^{l,j}}

%Arindam theorem
\renewcommand\thesubfigure{(\alph{subfigure})}
\theoremstyle{plain}
% \newtheorem{definition}{Definition}[section]
% \newtheorem{theorem}{Theorem}
% \newtheorem{lemma}{Lemma}

% Arindam eq spacing
\newcommand{\smalleq}[1]{
    \footnotesize
    #1
    \normalsize
}

\begin{abstract}
Selecting the right set of hyperparameters is crucial in time series forecasting. The classical temporal cross-validation framework for hyperparameter optimization (HPO) often leads to poor test performance because of a possible mismatch between validation and test periods. To address this test-validation mismatch, we propose a novel technique, \hpro~to drive HPO via test proxies by exploiting data hierarchies often associated with time series datasets. Since higher-level aggregated time series often show less irregularity and better predictability as compared to the lowest-level time series which can be sparse and intermittent, we optimize the hyperparameters of the lowest-level base-forecaster by leveraging the \textit{proxy forecasts} for the test period generated from the forecasters at higher levels. \hpro~ can be applied on any off-the-shelf machine learning model to perform HPO. We validate the efficacy of our technique with extensive empirical evaluation on five publicly available hierarchical forecasting datasets. Our approach outperforms existing state-of-the-art methods in \tour, \wiki, and \traffic~ datasets, and achieves competitive result in \tourL~ dataset, without any model-specific enhancements. Moreover, our method outperforms the winning method of the \mfive~ forecast accuracy competition.
\end{abstract}

% \begin{CCSXML}
% <ccs2012>
%    <concept>
%        <concept_id>10010405.10010481.10010487</concept_id>
%        <concept_desc>Applied computing~Forecasting</concept_desc>
%        <concept_significance>500</concept_significance>
%        </concept>
%    <concept>
%        <concept_id>10010147.10010257.10010339</concept_id>
%        <concept_desc>Computing methodologies~Cross-validation</concept_desc>
%        <concept_significance>500</concept_significance>
%        </concept>
%    <concept>
%        <concept_id>10010147.10010257.10010293.10010294</concept_id>
%        <concept_desc>Computing methodologies~Neural networks</concept_desc>
%        <concept_significance>500</concept_significance>
%        </concept>
%  </ccs2012>
% \end{CCSXML}

% \ccsdesc[500]{Applied computing~Forecasting}
% \ccsdesc[500]{Computing methodologies~Cross-validation}
% \ccsdesc[500]{Computing methodologies~Neural networks}


\maketitle

\section{Introduction}\label{sec:Introduction}
% \textbf{--Hierarchical forecasting and applications--}\\
Time series data is often associated with a hierarchy. For example, in the retail domain, daily sales of a certain product in a store constitute a product-level time series. Aggregating all product-level time series in the store at each time point gives a cumulative store-level series consisting of the daily sales of that particular store. Similarly, aggregated time series can be obtained at other levels like department, state, and country~\cite{makridakis2022m5}.
The notion of generating forecasts at every level of the data hierarchy is generally termed as ``hierarchical time series forecasting''~\cite{hyndman_book}, and it has been an active area of research in recent years (see Section~\ref{sec:Related work}).
%~\cite{athanasopoulos2009hierarchical,hyndman_book,rangapuram2021end,spiliotis2019improving,das2022top,ben2019regularized,gleason2020forecasting,mancuso2021machine,shiratori2020prediction,kamarthi2022profhit,paria2021hierarchically}.
Hierarchical forecasting generally requires \textit{coherent} forecasts at every level; \ie the forecast at an aggregated level should be the exact sum of the forecasts of its children nodes in an associated hierarchy tree.
%~\cite{hyndman_book}.
Accurate and coherent forecasts at different levels of the hierarchy ensure that consistent and correct business decisions are taken at different parts of an organization. 
% For example, in inventory management, purchasing is a higher-level decision, while distribution is a granular decision. Hence, the forecasts must be coherent so that there is no excess or shortage of inventory, both of which are costly.

\begin{figure}[t]
     \centering
     \begin{subfigure}[b]{0.49\columnwidth}
         \centering
         \includegraphics[width=\columnwidth]{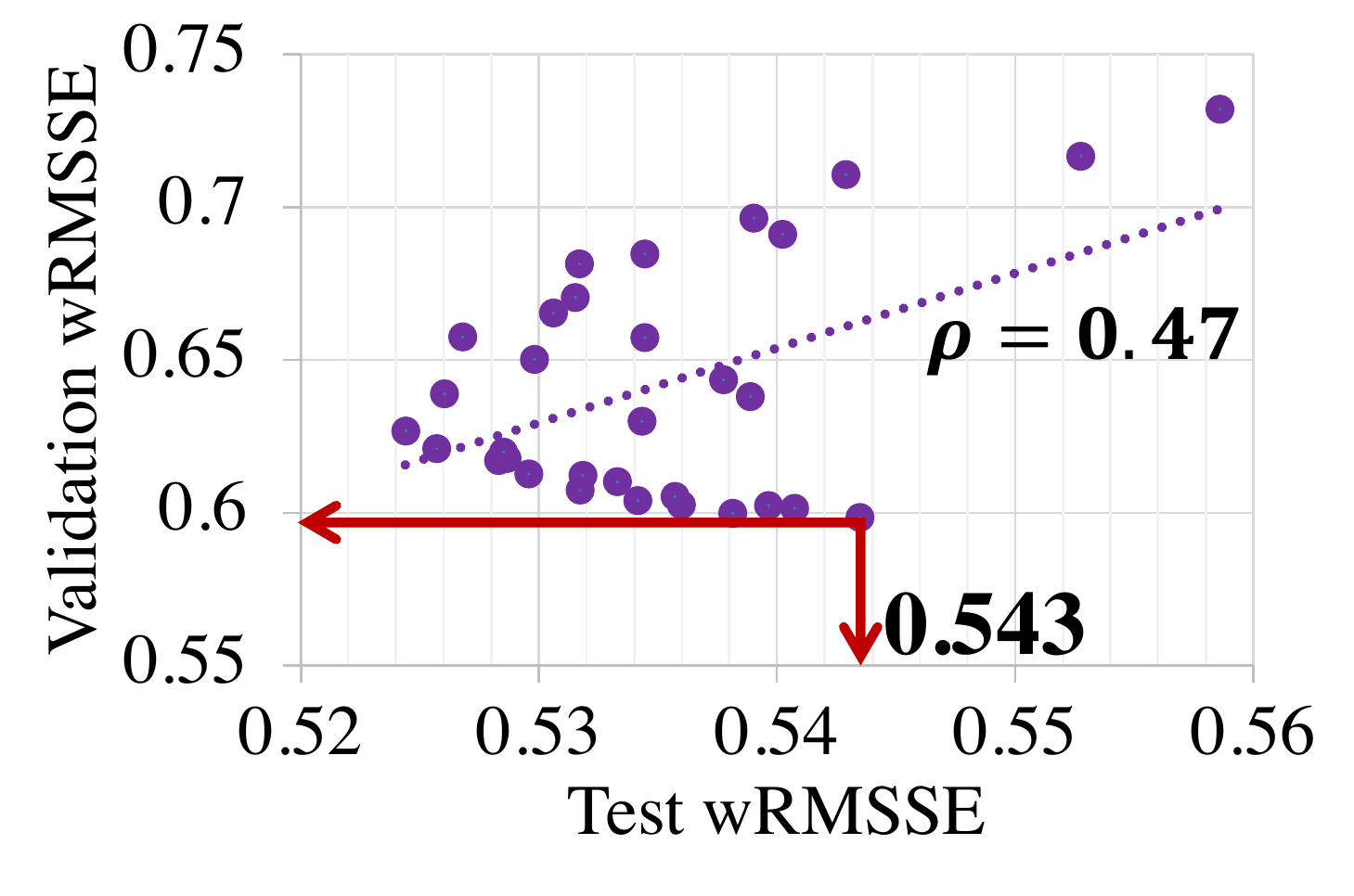}
         \caption{Temporal cross-validation}
         \label{fig:intro-motivation-heldout}
     \end{subfigure}
     \hfill
     \begin{subfigure}[b]{0.49\columnwidth}
         \centering
         \includegraphics[width=\columnwidth]{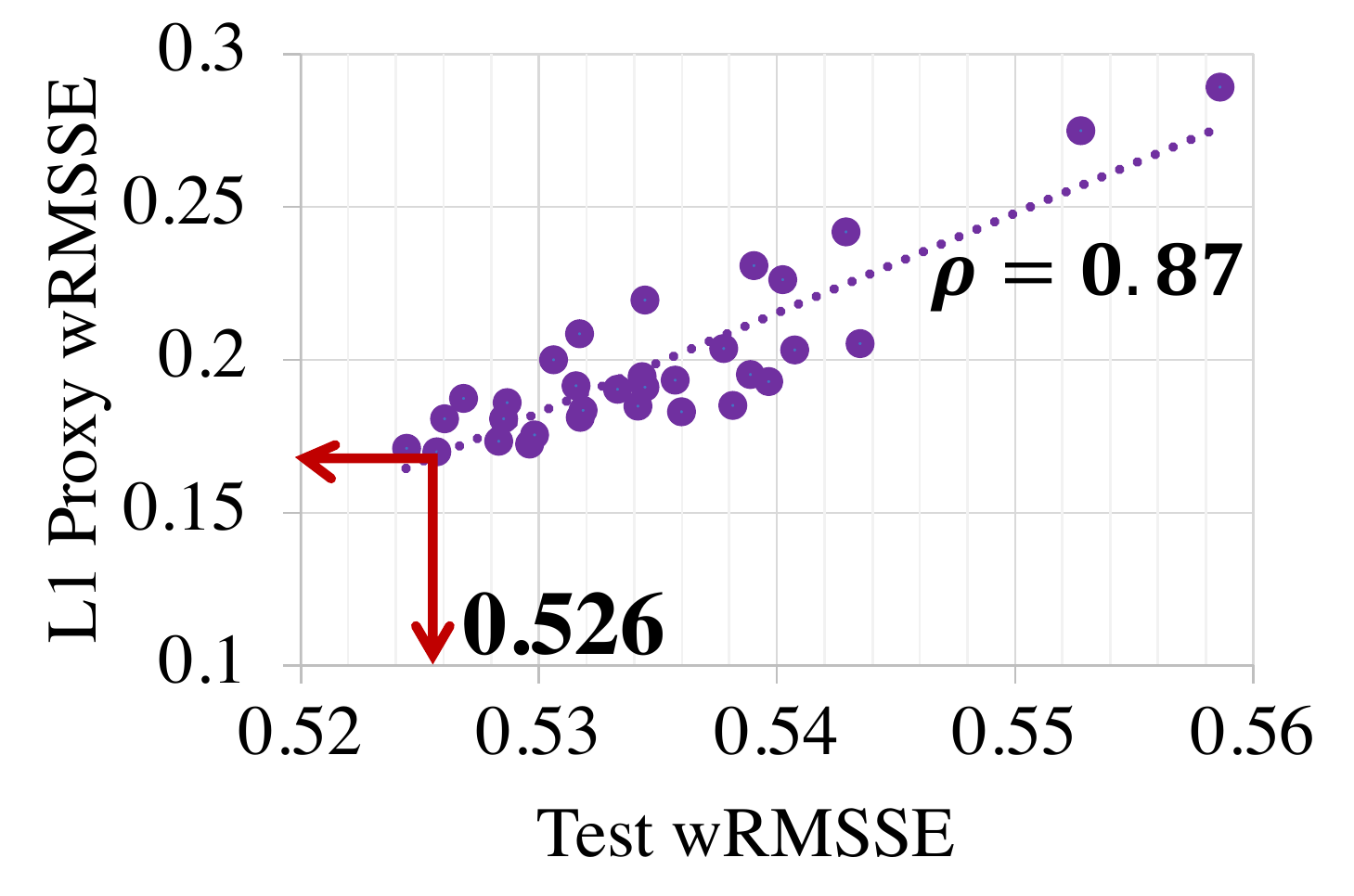}
         \caption{Proposed \hpro}
         \label{fig:intro-motivation-proxy}
     \end{subfigure}
    \caption{
    Variation of test error with (a) validation, and (b) proxy errors in all HPO trials (\textit{dots}) for a store-clustered LightGBM model on M5 data. 
    % The dots denote the HPO trials. 
    % In temporal cross-validation (a), the lowest validation error does \textit{not} correspond to lower range of test errors due to data mismatch. 
    (a): Lowest validation errors do \textit{not} correspond to the lower range of test errors due to data mismatch. 
    % In \hpro~(b), the lowest proxy error often selects model with better test errors.
    (b):  The lowest proxy error often selects a model with better test error, and better linear fit is observed with higher correlation, $\rho$, between test and proxy errors.
    % In (a), the lowest validation error does \textit{not} correspond to lower range of test errors due to data mismatch. 
    % % In \hpro~(b), the lowest proxy error often selects model with better test errors.
    % In (b), the lowest proxy error often selects models with better test errors, and better linear fit is observed with higher Pearson's correlation, $\rho$, between test and proxy errors.
    }
    \label{fig:intro_motivation_m5}
\end{figure}

% \textbf{--Popularity of complex models with lots of hyperparameters--}
% A hierarchical forecasting algorithm consists of two components (either decoupled or integrated into an end-to-end algorithm): 
A hierarchical forecasting algorithm consists of two components (either decoupled or integrated): 
a base-forecasting method, and a reconciliation technique that ensures coherent forecasts.
% ~\cite{hyndman_book, rangapuram2021end}.
% Recently, complex machine learning (ML) models with large number of parameters and hyperparameters are becoming popular in time series forecasting since they can learn from multiple time series and leverage the shared information between them, contrary to some of the classical forecasters like SARIMA, exponential smoothing, \etc~  
Recently, complex machine learning (ML) models with a large number of hyperparameters are becoming popular in forecasting since they can learn from multiple time series and leverage the shared information between them, contrary to some of the classical forecasters like SARIMA, exponential smoothing, \etc~  
Examples include gradient boosting models like LightGBM~\cite{makridakis2022m5}, and DNN-based models like DeepAR~\cite{salinas2020deepar}, N-BEATS~\cite{oreshkin2019n}, and Informer~\cite{zhou2021informer}.
This leads to an increase in the adoption of complex ML models for hierarchical forecasting as well due to their superior performance~\cite{rangapuram2021end,paria2021hierarchically,das2022top,mancuso2021machine}.

% \textbf{--Problem with model selection in (hierarchical) forecasting--}\\
% The performance of these models depends on hyperparameters that are generally tuned on one or more validation sets via temporal cross-validation (\tcv).
% The performance of these models depends on hyperparameters that are generally tuned on one or more validation windows via temporal cross-validation~(\tcv)~\cite{hyndman_book}.
The performance of these models depends on hyperparameters that are generally tuned on validation window(s) via temporal cross-validation~(\tcv)~\cite{hyndman_book}.
\tcv~chronologically splits the historical time series data into train and validation windows.
Thus, the validation windows often differ in characteristics from the test data.
% The validation periods are chosen from the historical data, and thus, often differ in characteristics from the actual test data.
% The mismatch between validation and test is prevalent in time series compared to other ML domains because of the highly non-stationary and sometimes heteroscedastic nature of temporal data~\cite{hyndman_book,akgiray1989conditional}.
The mismatch between validation and test is prevalent in time series compared to other ML domains because of varying statistical properties of temporal data~\cite{hyndman_book,akgiray1989conditional}.
The problem is exacerbated in hierarchical forecasting because of higher data irregularity (and sometimes intermittency) at the lowest level.
This can lead to suboptimal hyperparameter optimization (HPO) and poor model selection, particularly at the lowest level of the hierarchy (see Figure~\ref{fig:intro-motivation-heldout}).
% Figure~\ref{fig:intro-motivation-heldout} demonstrates this issue for the M5 dataset.

% \textbf{--proposed method, how hierarchy can be exploited for model selection--}\\
We propose a novel HPO technique for hierarchical time series forecasting.
It is based on the frequent observation that time series at the lowest level of the hierarchy are sparse, irregular, and sometimes intermittent in nature such as in~\cite{makridakis2022m5}. 
However, the aggregated series at higher levels are generally more consistent and have better predictability. 
% Figure~\ref{fig:sparsity_explain} shows this pattern for M5 dataset.
Based on this observation, we develop \hpro, a method for performing HPO of the lowest-level forecasting model based on one or more \textit{proxy forecaster(s)} at higher levels.
% of the hierarchy.
The proxy forecasters are trained on higher-level aggregated time series, and their forecasts for the \textit{test} period are obtained.
The lowest-level model treats these forecasts as proxies to the original time series for the test period, and the HPO of the lowest-level model is performed with respect to the higher-level proxy forecasts instead of validation windows as done in conventional \tcv~(Figure.~\ref{fig:proxy_method}).
Thus, by effectively leveraging the better predictability of the aggregated series, the lowest-level models are regularized via HPO.
The lowest-level forecasts are then aggregated bottom-up (Section~\ref{sec:Related work}) to derive higher-level forecasts leading to coherent and accurate forecasts at all levels.
From Figure~\ref{fig:intro-motivation-proxy}, we can see that \hpro~helps in model selection, \ie choosing a model with the minimum proxy error criteria corresponds to a much better test error compared to conventional \tcv. 

% \textcolor{orange}{TODO: Can we add one line detailing the bottom up aggregation of the lowest-level forecasts till the hierarchy level to derive the proxy error metric for a hierarchy level. In Figure~\ref{fig:intro-motivation-proxy}, we refer to BU aggregated forecast, but the details of bottom-up forecast aggregation and its associated abbreviation(BU) is missing.}

% ----- deprecated ----------
% \begin{figure}[t]
%     \centering
%     \includegraphics[width=\columnwidth]{figures/sparsity_explain.pdf}
%     \caption{Sparse and incoherent time series at the lowest level of M5 dataset. At the higher levels, we observe better coherence of the data. The time series are randomly chosen from the M5 dataset from the respective levels. Toy icons are used at the nodes for illustration purpose only.}
%     \label{fig:sparsity_explain}
% \end{figure}
% ------- end deprecated ----------

% \textbf{contributions}\\

\subsubsection{Summary of contributions}
\begin{enumerate*}[(1)]
    \item To address the commonly occurring test-validation mismatch issue in forecasting, we propose a novel technique, \hpro~that drives HPO via test proxies by exploiting the data hierarchy and better predictability of higher level forecasters.
    
    \item To the best of our knowledge, this is the first work which empirically and theoretically demonstrate that we can obtain coherent and accurate hierarchical forecasts just by employing hierarchical proxy-guided HPO on off-the-shelf ML models. Specifically, \hpro~outperforms state-of-the-art results in \tour, \traffic, and \wiki~data, and achieves competitive result in \tourL~ data. Moreover, \hpro~ outperforms the winning method of the \mfive~ forecast accuracy competition.
    
    % \item Moreover, we provide theoretical analysis that establishes the characteristics of the proposed approach.
    
    % \item To the best of our knowledge, this is the first work to demonstrate that we can obtain coherent and accurate hierarchical forecasts by employing \hpro~based HPO on off-the-shelf ML models. Specifically, \hpro~outperforms state-of-the-arts in \tour, \tourL, \traffic, and \wiki~data, and the winning algorithm of \mfive~competition.
    
    % \item \hpro~is a novel HPO technique that leverages the better predictability of a set of models to regularize another model by exploiting the data hierarchy.
    
    \item State-of-the-art hierarchical reconciliation methods like \mint~and \erm~(Section~\ref{sec:Related work}) can be computationally expensive for datasets with large number of time series, but the proposed \hpro~does not suffer from this scalability issue. A detailed comparison is in Appendix~A.
    
    % \item Moreover, we provide theoretical analysis that establishes the characteristics of the proposed approach.
    % \item \hpro~has similar or better computational need as compared to \tcv~with single or multiple validation windows respectively.
    \item We also show in experiments that \hpro~helps improve the performance of \tcv~when ensembled with it, and hence, emphasize the complementary knowledge captured by the method.
\end{enumerate*}

% \hpro~comes with the following benefits. It has no restriction on the hierarchy tree-structure, and works smoothly with naive bottom-up aggregation (Section~\ref{sec:Background}).

% \hpro~outperforms state-of-the-art methods in several datasets with off-the-shelf ML models.
% \textcolor{orange}{Can we have a seperate key contribution section summarising and highlighting the novelty of the approach (first-of-the-kind, etc.), why this approach is different as compared to existing reconciliation methods,  empirical analysis (highlighting improvements wrt. SOTA and beating M5), theoretical analysis and benefits of our approach wrt. runtime and performance ? }
% ------- OLD
% \begin{itemize}
%     \item It works on single as well as grouped hierarchies~\cite{hyndman_book}, always produces coherent forecasts.  \textcolor{red}{do not follow what is a grouped hierarchy}
%     \item It does not need any validation set or temporal cross-validation to perform the HPO of the lowest-level model. Thus, the runtime for HPO can be significantly reduced in comparison with temporal cross-validation with multiple validation sets.
%     \item It outperforms several baselines and state-of-the-art algorithms in multiple hierarchical forecasting datasets.
% \end{itemize}
%-------- END OLD

\section{Background and Notations}\label{sec:Background}
\subsection{Forecasting}\label{subsec:Forecasting}
Let $x_{1:T}$ represent a univariate time series of length $T$, \ie at any time-point $t$, $x_t \in \mathbb{R}$.
% At any time point $t$, the time series takes a scalar value, \ie $x_t \in \mathbb{R}$. 
The task of forecasting is to predict $H$ value(s) in the future given the history $x_{1:T}$,
\begin{equation}\label{eq:forecast_def}
    \hat{x}_{T+1:T+H} = f(x_{1:T})
\end{equation}
where $f(\cdot)$ is the map learned by a forecasting algorithm $\mathcal{A}$, and $H$ is the forecast horizon.
% $H$ is known as the ``forecast horizon''.
Here we focus on algorithms that learn a shared map $f$ for multiple related time series, as these often work best in practice (\eg modern forecasting models like DeepAR, N-BEATS, and LightGBM). 
% In this work, we focus on algorithms, $\mathcal{A}$ that can learn the map $f$ from multiple time series together (\eg the modern forecasting models like DeepAR, N-BEATS, and LightGBM). 

% --------- This is not being used anywhere in the methodology, this is relevant for M5 feature processing, but we do not have space to describe that in main text. We can have this in the appendix.
% In some domains (\eg M5 dataset), the value of the time series also depends on few other related time series known as ``external regressors'' (or factors).
% \begin{equation}
%     \hat{x}_{T+1:T+H} = f(x_{1:T}, \{y^{i}_{1:T}\}_{i=1}^{n_y}, \{z^{i}_{1:T+H}\}_{i=1}^{n_z})
% \end{equation}
% Here, $y^{i}_{1:T}$ is an external regressor time series available only in the history (such as past natural calamities can affect the sales of a product). On the other hand, $z^{i}_{1:T+H}$ denotes an external regressor available for both past and future (such as the discount of a product which is generally decided for the forecast horizon as well).

\subsection{Hierarchical forecasting}\label{subsec:Hierarchical forecasting}
A hierarchical time series dataset is associated with a hierarchy tree that has $L$ levels ($l=1$ for the top level, and $l=L$ for the lowest level).
We denote the set of levels by $[L]=\{1,\dots,L\}$, and the nodes at a level $l$ by $[N_l]=\{1,\dots,N_l\}$. 
The time series at the leaf nodes are called \textit{lowest-level} series, and the time series at other nodes are called \textit{higher-level/aggregated} series.
The higher-level series at any node $j\in [N_l]$ of level $l \in [L-1]$ follows the coherence criteria \cite{hyndman_book}: $\xlj = \sum_{i \in {C_{l,j}}} x^{(l+1),i}_t$, where $C_{l,j}$ is the set of children of 
% the particular node in concern.
node $j$ of level $l$.
The task is to forecast accurately at all nodes of all levels so that the forecast at any higher-level node also follows the coherence constraint, \ie $\xhatlj = \sum_{i \in {C_{l,j}}} \hat{x}^{(l+1),i}_t, \forall l \in [L-1], j\in [N_l]$.

\subsubsection{Hierarchical evaluation}
Hierarchical forecasts are evaluated with a hierarchically aggregated metric that can indicate the average error across all levels~\cite{makridakis2022m5}.
Generally, every level is given equal weight while aggregating the level-wise metrics, ensuring unbiased evaluation of hierarchical forecasts across all levels. 
Hence, any hierarchical forecasting technique should aim to attain accurate and coherent forecasting at all levels.
More details will be provided in Section~\ref{sec:Experiments}.

%%% Ignoring the external regressors for simplicity, the forecast at any node $j\in [N_l]$ at level $l \in [L]$ is given by
% The forecast at any node $j\in [N_l]$ at level $l \in [L]$ is given by
% \begin{align}\label{eq:hier_forecast}
%     \hat{x}^{l,j}_{T+1:T+H} &= f^{l,j}(x^{l,j}_{1:T})
%     \quad \mathrm{s.t.} \quad \hat{x}^{l,j}_t  &= \sum_{i=1}^{C_{l,j}} \hat{x}_t^{(l+1),i}
% \end{align}
% where the constraint is \textit{not} applicable for the lowest level. \textcolor{red}{Eq not very clear and maybe can be made more crisp}
% Here, $C_{l,j}$ is the number of children of the particular node in concern, and $f^{l,j}(\cdot)$ is the corresponding forecaster model.

% {\bf Comment:} Do we want to add something to the definition to differentiate the forecast at the $j$th node in the $l$th level, so that $\mathrm{children}$ can be precise?

% While numerous algorithms of hierarchical forecasting exists in the literature (as will be explained in Section~\ref{sec:Related work}), the bottom-up (BU) aggregation technique will be employed in the proposed method. It obtains forecasts at the lowest level, and simply aggregates them to obtain forecaster at any higher level~\cite{hyndman_book}.

\section{Related work}\label{sec:Related work}
Classical methods of hierarchical forecasting rely on generating base forecasts for every time series, and reconcile them to produce coherent forecasts at every level.
For example, the bottom-up (\bu) approach produces lowest level forecasts, and simply aggregates them to obtain coherent forecasts at all levels~\cite{hyndman_book}. 
Similarly, top-down and middle-out approaches take particular aggregate level forecasts and disaggregate them to lower levels \cite{hyndman_book}.
% Similarly, top-down and middle-out approaches also exist as described in \cite{hyndman_book}.
The \mint~ optimal reconciliation algorithm takes independent forecasts and produces coherent hierarchical forecasts by incorporating information from all levels simultaneously via a linear mapping to the base series~\cite{wickramasuriya2019optimal, hyndman_book}. 
\mint~minimizes the sum of variances of the forecast errors when the individual forecasts are unbiased. \cite{ben2019regularized} relaxed the unbiasedness condition, and proposed \erm~which optimizes the bias-variance trade-off by solving an empirical risk minimization problem.
Since \erm~ is a successor of \mint, we refer both of them as optimal reconciliation algorithms throughout the text.
\cite{rangapuram2021end} proposed \hieretoe~which trains a single neural network on all time series together.
It enforces coherence conditions in model training.
Another end-to-end approach was proposed in \cite{mancuso2021machine} where the reconciliation is imposed in a customized loss function of the neural network.
A probabilistic top-down approach was proposed in \cite{das2022top} where a distribution of proportions is learnt by an RNN model to split the parent forecast among its children nodes.
A top-down alignment-based reconciliation was developed in \cite{anderer2021forecasting} where the lowest-level forecasts are adjusted based on the higher-level forecasts. The method employs a bias-controlling multiplier for the loss function of the lowest-level model, 
optimized by manual grid search. 
% which was optimized by manual grid search. 
We want to highlight that the proposed \hpro~is compatible with any search algorithm (see Section~\ref{subsec:HPO framework} for details). Moreover, AutoML frameworks like~\cite{abdallah2022autoforecast, deng2023efficient} can internally employ H-Pro for hierarchical forecasting tasks.

\section{\hpro} \label{sec:HPRO}
\begin{figure}[t]
    \centering
    \includegraphics[width=\columnwidth]{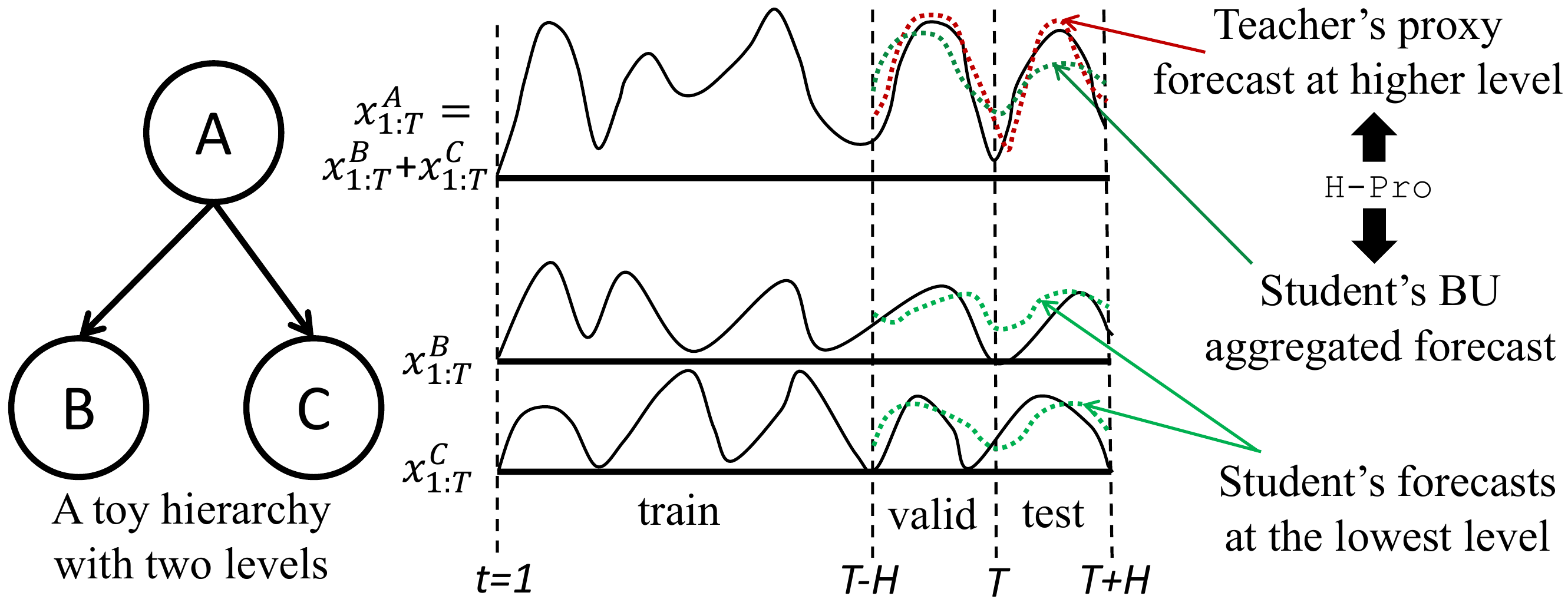}
    \caption{Visual explanation of \hpro~for a toy hierarchy. Temporal cross-validation performs HPO on the validation period with the actual ground truth data. \hpro~performs HPO on the test period with the teacher's proxy forecast. 
    %Note that $x^A_t=x^B_t+x^C_t$, but this might not appear visually since toy time series are shown for illustration purpose only.
    }
    \label{fig:proxy_method}
\end{figure}
% We demonstrate how HPO is performed for a complex ML model with conventional temporal cross-validation. Then, we derive the same for the proposed \hpro~method.
We derive the HPO objective for conventional \tcv, and then, extend it for \hpro.
For both cases, the models are trained at the lowest level, but their HPO techniques differ.
Bottom-up (\bu) aggregation is employed in both scenarios to generate coherent forecasts.
% A learning algorithm generally has two optimization procedures: inner and outer.
% The parameters, $\theta$ of the algorithm are learned in the inner optimization, while the hyperparameters, $\lambda$ are learned in a (generally more complex) outer optimization.
% Ignoring the $\theta$ parameters (since that is not our focus), the algorithm can be formally denoted as $\algo{train}$, where $\dset{L}{train}$ is the training dataset at the lowest level.
We denote a learning algorithm by $\algo{train}$, where $\dset{L}{train}$ is the training data at the lowest level, $\lambda$ denotes the hyperparameters, and we ignore the model's parameters since those are learned in a separate optimization regime (not our focus).
The goal of HPO is to find the best $\lambda$ by minimizing an objective.
% For a given $\lambda$, $\mathcal{A}$ provides the forecasting map $f_\lambda: \mathbb{R}^{T} \rightarrow \mathbb{R}^H$ as shown in \eqref{eq:forecast_def}.
For a given $\lambda$, $\mathcal{A}$ provides the forecasting map $f_\lambda$ as shown in \eqref{eq:forecast_def}.
% ------- deprecated, not a good representation -------
% \begin{equation}
%     f_\lambda = \mathcal{A}\left( \mathcal{X}^L_{\mathrm{train}}; \lambda \right)
%     \label{eq:A}
% \end{equation}
% ------------- end deprecated ------------------------
We subscript $f$ with $\lambda$ to concisely denote the forecasting model's dependency on the algorithm's hyperparameters. 
% --- OLD ------------
% \begin{equation}
%     \hat{x}^L = f_\lambda \left( x^L \right) \quad \forall x^L \in \mathcal{X}^L_{\mathrm{train}}
% \end{equation}
% Generally, the immediate past is utilized by the model to forecast for a certain period. Hence, forecasts for validation and test periods are given by (see Figure~\ref{fig:proxy_method}),
% \begin{subequations}
%     \begin{equation}
%         \hat{x}^{\left( \text{valid, L} \right)} = f\left( x^{\left( \text{train, L} \right)} \right)
%         \label{eq:forecast-valid}
%     \end{equation}
%     \begin{equation}
%         \hat{x}^{\left( \text{test, L} \right)} = f\left( x^{\left( \text{train+valid, L} \right)} \right)
%         \label{eq:forecast-test}
%     \end{equation}
% \end{subequations}
% \begin{equation}
%     \hat{x}^{\left( \text{valid, L} \right)} = f\left( x^{\left( \text{train, L} \right)} \right), \quad
%     \hat{x}^{\left( \text{test, L} \right)} = f\left( x^{\left( \text{train+valid, L} \right)} \right)
%     \label{eq:valid_train_forecast}
% \end{equation}
% {\bf comment:} Should we distinguish the two $f$ above? Hyperparameters are identical, but internal parameters are different.
% --- END OLD ------------

\subsection{HPO with temporal cross-validation (TCV)}\label{subsec:HPO with temporal cross-validation (TCV)}
Traditionally, \tcv~has been used to  perform HPO and model selection for forecasting~\cite{salinas2020deepar,oreshkin2019n}.
Figure~\ref{fig:proxy_method} shows the train, validation, and test splits for \tcv~with one validation window.
% Generally, test window is provided in a dataset (often it is the last forecast horizon). The rest of the history constructs the train+validation data. The validation window(s) are extracted from the data before the test windows.
Hence, those subsets can be expressed with time-ranges. For example, $\dset{L}{train} = \{ x^{L,j}_{1:T-H} \}_{j=1}^{N_L}$, and so on.
% Here, $N_L$ denotes the total number of time series at the lowest level of the hierarchy (\ie number of leaf nodes).
% Note that here the training is done at the lowest level of the hierarchy, $L$.
We show the HPO objectives below for one validation window, but it can be generalized to multiple windows as well.
Irrespective of the number of validation windows, the model is trained again on the entire train and validation period (\ie on $\dset{L}{train+valid}$) with the optimal hyperparameters $\lambda^*$ to produce test forecasts, $\hat{x}^{L,j}_{T+1:T+H}$. 
% Hence, following the notations from Section~\ref{sec:Background}, the validation and test forecasts can be expressed as 
% \begin{equation}\label{eq:forecast-valid-test}
%     \hat{x}^{L,j}_{T-H+1:T} = f_\lambda \left( x^{L,j}_{1:T-H} \right),
%     \hat{x}^{L,j}_{T+1:T+H} = f^{\prime}_\lambda \left( x^{L,j}_{1:T} \right)
% \end{equation}
% where, $f^{\prime}_\lambda$ is the model retrained on $\dset{L}{train+valid}$.

The HPO for \tcv~can target either the lowest-level error, or a hierarchically aggregated error (see Section~\ref{subsec:Hierarchical forecasting}). We denote the two variants as \tcv-\lowest~and \tcv-\hier.
% Following the notations of \cite{bergstra2012random}, the HPO objective for a single-validation \tcv-\lowest~can be written as
% \begin{equation}
%     \lambda^\ast_{\mathrm{TCV\emph{-}Lowest}} = \argmin_{\lambda \in \Lambda} \mathbb{E}_{x^L \sim G_{x^L}} \left[ \mathcal{L}\left( x^L, \hat{x}^L \right) \right]
%     \label{eq:hpo-obj-backtest-true}
% \end{equation}
% where, $G_{x^L}$ is the true natural distribution, and $\mathcal{L}$ is the HPO objective. 
% Since, $G_{x^L}$ is unknown, we approximate \eqref{eq:hpo-obj-backtest-true} with the following.
Following \cite{bergstra2012random}, the HPO objective for a single-validation \tcv-\lowest~can be written as
\begin{align} \label{eq:hpo-obj-backtest-approx}
    & \lambda^\ast_{\mathrm{TCV\emph{-}Lowest}}  \approx \argmin_{\lambda \in \Lambda} 
    \mean_{x^L \in \mathcal{X}^L_{\mathrm{valid}}}
    \left[ \mathcal{L}\left( x^L, \hat{x}^L \right) \right] \nonumber \\
   & = \argmin_{\lambda \in \Lambda} \frac{1}{N_L} \sum_{j=1}^{N_L} \mathcal{L} \left( 
    x^{L,j}_{T-H+1:T}, 
    f_\lambda \left( x^{L,j}_{1:T-H} \right) 
    \right).
\end{align}

% While \tcv-\lowest~focuses on regularizing the model for better lowest-level error, \tcv-\hier~can also be beneficial since it tries to obtain better error at all levels. Formally,
While \tcv-\lowest~ targets minimizing lowest-level error, \tcv- \hier~ better targets the hierarchical forecasting objective as it aims to obtain low error at all levels. Formally,
\begin{align}\label{eq:TCV-Hier}
    &\lambda^\ast_{\mathrm{TCV\emph{-}Hier}} 
    \approx \argmin_{\lambda \in \Lambda} 
    \frac{1}{L} \sum_{l=1}^{L} \biggl[ \frac{1}{N_l} \times \biggl. \nonumber \\
    & \sum_{j=1}^{N_l} 
    \left[ \mathcal{L} \left( x^{l,j}_{T-H+1:T}, 
    \mathcal{B} \left( f_\lambda, \left\{ x^{L,i}_{1:T-H} \right\}_{i=1}^{N_L}, l, j \right)  
    \right) \right]
    \biggr. \biggr]
\end{align}
where, $\mathcal{B}\left(\cdot\right)$ is the bottom-up aggregation function which aggregates forecasts from the descendant leaf nodes to produce a forecast at node $j$ of level $l$.

% {\bf Comment:} In (9) $\mathcal{A}$ returns a function\ldots I think we need $\mathcal{A}_\lambda \left( \mathcal{X}^{\left( \text{train, L} \right)} \right)(x^{(\text{train, L})})$, similar comment for (10)
% \textcolor{orange}{TODO: while reading, it looks like we are suddenly jumping from lowest-level metric in TCV to hierarchical metric in H-PRO. It could be bit confusing to readers on this sudden switch in the used metric. So - can we define TCV-Hier also in this section where we motivate the need to minimize the hierarchical metric as opposed to lowest metric to get coherent and accurate forecasts at all levels as 
% compared to the lowest level. Once we define the hierarchical metric, then H-PRO section continues to use it. Since lowest-level metric is the commonly used metric, it is good to explain the importance of hierarchical metric before in hand. Else, while reading the theorems, where
% we talk about improving higher level predictions, reader can get confused on how it can improve the lowest metric.}

\subsection{HPO with hierarchical proxy modeling}\label{subsec:HPO with hierarchial proxy modeling}
From \eqref{eq:hpo-obj-backtest-approx} and \eqref{eq:TCV-Hier} we can see that $\lambda^\ast_{\mathrm{TCV\emph{-}Hier}}$ and $\lambda^\ast_{\mathrm{TCV\emph{-}Lowest}}$ depend on the validation data $x^{l,j}_{T-H+1:T}$.
%and the training data $\dset{train}$ at the lowest level.
Hence, a mismatch between the test series $x^{l,j}_{T+1:T+H}$ and validation series $x^{l,j}_{T-H+1:T}$ can lead to poor test performance.
\hpro~attempts to address the issue based on the observation that, often, the higher-level aggregated time series are less irregular and have better predictability (\eg in \cite{makridakis2022m5}).
It builds two sets of models: a \textit{student model} as a base forecaster at the lowest level, and \textit{teacher model(s)} as proxy forecasters at any (or all) higher levels (see Figure~\ref{fig:proxy_method})\footnote{\hpro~is different from knowledge distillation~\cite{hinton2015distilling}. \hpro~is developed for HPO and not for model training, it requires a hierarchical dataset, and it differs in the core algorithm.}.
The student produces the final forecasts at all levels via bottom-up (\bu) aggregation.

\hpro~proceeds as follows. 
The teacher model(s) are trained and their HPO is performed with \tcv.~Teacher's forecasts are generated for the \textit{test} period. We term these as teacher's \textit{proxy forecasts} since the student treats them as the actual ground truth for the \textit{unknown} test period.
The student is trained on the entire train and validation data at the lowest level, but its HPO is performed based on the proxy forecasts at higher levels.
Intuitively, the student model is regularized in a way such that it tries to mimic the proxy forecasts of the teacher, but only at higher level(s) since teacher's forecasts are \textit{not} available at the lowest level.
We hypothesize that even if we guide the student to produce accurate forecasts at higher level(s) for the test period, it would enable the student to produce accurate forecasts in all or at least some of the lower levels because the higher level forecasts are obtained by aggregating the lower level forecasts (Figure.~\ref{fig:proxy_method}).

% ----------- OLD NOT NEEDED ---------------------
% Following \eqref{eq:hpo-obj-backtest-approx1}, the HPO objective of \hpro~can be written as 
% \begin{equation}
%     \lambda^\ast_{\texttt{H-Pro}}
%     \approx \argmin_{\lambda \in \Lambda} 
%     \sum_{l=1}^{L-1} \biggl[ w(l) \times \biggl.
%     \mean_{ \Tilde{x}^l \in \dset{l}{proxy} } 
%     \left[ \mathcal{L} \left( \Tilde{x}^l, \hat{x}^l \right) \right] 
%     \biggl. \biggr]
%     \label{eq:hpo-obj-proxy-approx1} 
% \end{equation}
% \textcolor{red}{why $\Tilde{x}^l \in \dset{l}{proxy} $, maybe the definition below will suffice}
% ----------- END OLD ---------------------------
Following \eqref{eq:TCV-Hier}, the \hpro~objective can be written as 
\begin{align}
    &\lambda^\ast_{\texttt{H-Pro}}
    \approx \argmin_{\lambda \in \Lambda} 
    \sum_{l=1}^{L-1} \biggl[ w(l) \cdot \frac{1}{N_l} \times \biggl. \nonumber \\
    & \sum_{j=1}^{N_l} 
    \left[ \mathcal{L} \left( \Tilde{x}^{l,j}_{T+1:T+H}, 
    \mathcal{B} \left( f_\lambda, \left\{ x^{L,i}_{1:T} \right\}_{i=1}^{N_L}, l,j \right) 
    \right) \right]
    \biggr. \biggr]
    \label{eq:hpo-obj-proxy-approx2} 
\end{align}
where, $\Tilde{x}^{l,j}_{T+1:T+H}$ denotes the teacher's proxy forecasts at that node, and $w(l) \in [0,1]$ assigns a confidence-weight on the teacher at level $l$. Hence, $\sum_{l=1}^{L-1}w(l)=1$.
%\footnote{Usage of fractional weights is left to further studies.} 
It is evident that the optimal hyperparameters $\lambda^\ast_{\texttt{H-Pro}}$ depend on the teacher's proxy forecasts $\Tilde{x}^{l,j}_{T+1:T+H}$, and the bottom-up aggregated test forecast of the student.
%, where the latter is obtained through the student model trained on the entire train and validation period, $\mathcal{X}^{L}_{1:T}$. 
This removes the dependency of \hpro-based HPO on the validation period.

\subsubsection{Properties of \hpro} We describe some characteristics of \hpro~which will help us understand its strength. %as a hierarchical HPO technique.
\begin{definition}[Perfect teacher]\label{def:Perfect teacher}
A perfect teacher generates proxy forecasts with zero error, \ie $\forall l \in [L-1], j \in [N_l]$, $\Tilde{x}^{l,j}_{T+1:T+H} = x^{l,j}_{T+1:T+H}.$
% \begin{equation} \label{eq:perfect-teacher}
%     \Tilde{x}^{l,j}_{T+1:T+H} = x^{l,j}_{T+1:T+H}.
% \end{equation}
\end{definition}

\begin{definition}[\texttt{OPT-BU}]
The optimal bottom-up (\texttt{BU})-aggregated student model is obtained by optimizing the hyperparameters of a student model, where the HPO objective minimizes a specified loss $\mathcal{L}$ between the bottom-up aggregated student forecasts and the ground truth data at the test period across all higher levels.
\emph{
\begin{align}\label{eq:OPT-BU}
    &\lambda^\ast_{\texttt{OPT-BU}} 
    \approx \argmin_{\lambda \in \Lambda} 
    \frac{1}{L-1} \sum_{l=1}^{L-1} \biggl[ \frac{1}{N_l} \times \biggl. \nonumber \\
    & \sum_{j=1}^{N_l} 
    \left[ \mathcal{L} \left( x^{l,j}_{T+1:T+H}, 
    \mathcal{B} \left( f_\lambda, \left\{ x^{L,i}_{1:T} \right\}_{i=1}^{N_L}, l,j \right)  
    \right) \right]
    \biggr. \biggr].
\end{align}
}
\end{definition}
\begin{lemma}\label{th:1}
    For a perfect teacher, if \emph{$w(l)=\frac{1}{L-1}, \forall l \in [L-1]$} in \eqref{eq:hpo-obj-proxy-approx2}, the hyperparameters of the student model obtained by applying \emph{\hpro}~are the same as that of the \emph{\texttt{OPT-BU}}, \ie \emph{$\lambda^\ast_{\texttt{H-Pro}} = \lambda^\ast_{\texttt{OPT-BU}} $}.
\end{lemma}
% \begin{proof}
%     Using Def.~\ref{def:Perfect teacher} in \eqref{eq:hpo-obj-proxy-approx2}, and comparing with \eqref{eq:OPT-BU}.
% \end{proof}
The proof is in Appendix~\ref{app:B}. Lemma~\ref{th:1} implies that if the teacher is extremely accurate, \hpro~can regularize the student to have accurate aggregated forecasts at the higher levels. However, a perfect teacher is rare! The following theorem attempts to quantify the difference between the HPO objectives of \hpro~with an imperfect and a perfect teacher.

\begin{theorem}\label{th:2}
    Let \emph{$\epsilon_t^{l,j} = |x_t^{l,j} - \Tilde{x}_t^{l,j}|$} and \emph{$\delta_t^{l,j} = |x_t^{l,j} - \hat{x}_t^{l,j}|$} be point-wise absolute errors for the teacher and the \texttt{BU}-aggregated student forecasts at the $j$-th node of $l$-th level. Let \emph{$\mathcal{E}$} denote the teacher's aggregated mean squared error at all higher levels. Let $\mathcal{O}$ and $\mathcal{O}^\ast$ denote the HPO objectives of \emph{\hpro}~and \emph{\texttt{OPT-BU}} respectively. 
    Let $w(l)=\frac{1}{L-1}, \forall l \in [L-1]$.
    Then, for mean squared error objective \emph{$\mathcal{L}$},
    \emph{
\begin{equation}\label{eq:th:2}
    \left| \mathcal{O} - \mathcal{O}^\ast \right|
    \le \mathcal{E} + \frac{2}{L-1} \sum_{l=1}^{L-1}\frac{1}{N_l} \sum_{j=1}^{N_l} \frac{1}{H} \sum_{t=T+1}^{T+H}
    \epsilon_t^{l,j} \delta_t^{l,j}
\end{equation}
    }
\end{theorem}
% \begin{proof}
%     The proof is provided in the appendix.
% \end{proof}

The proof is in Appendix~\ref{app:B}. The second term in the RHS of \eqref{eq:th:2} indicates an inter-related absolute error between the teacher and the \texttt{BU}-aggregated student forecasts.
The significance of Theorem~\ref{th:2} is that if we can a have reasonably accurate teacher ($\epsilon_t^{l,j} \rightarrow 0 \implies \mathcal{E} \rightarrow 0$), then, \hpro's objective is close to that of \texttt{OPT-BU}, leading to accurate forecasts at the higher levels.
On the contrary, a suboptimal teacher can lead to inferior performance of the student.
However, as explained above, the higher level time series generally possess better predictability (see Appendix~\ref{app:Example of reduced variance at higher level} for an example), and we observe this phenomena in multiple datasets, which leads to superior performance of \hpro~in our extensive experiments (Section~\ref{sec:Experiments}).

% One notable point is that \hpro~cannot guarantee accuracy of the student at the lowest level, but we validate our hypothesis as stated above through extensive experimentation.
One notable point is that the student is \textit{not} trained with proxy forecasts, but they are only regularized with them. 
The training is performed on the lowest-level's ground truth data from the entire train and validation periods. 
Hence, \hpro~does not have any direct effect on the learned parameters of the student model, but only on its hyperparameters. 

% \textcolor{red}{Tourism-LG student is an ensemble of theta and lightgbm. Table needs update}

\section{Experiments}\label{sec:Experiments}
\subsection{Experimental setting}\label{subsec:Experimental setting}
\subsubsection{Datasets}\label{subsec:Datasets}
\begin{table}[t]
    \setlength{\tabcolsep}{2.5pt}
    \centering
    \caption{Datasets and models. $N_L$= Number of lowest-level series, $H$= forecast horizon, $T+H$= length of series, $L$= Number of hierarchy levels.}
    \begin{tabular}{c c c c c | c c} 
        \toprule
        Dataset & $N_L$ & $T$ & $H$ & $L$ & Teacher & Student \\
        \midrule
        \tour & 56 & 28 & 8 & 4 & \deepar & \deepar \\
        \tourL & 304 & 216 & 12 & 8 & \thetamodel & \makecell{(\thetamodel+\lgbm)}\\
        \wiki & 150 & 365 & 1 & 5 & \deepar & \deepar\\
        \traffic & 200 & 359 & 7 & 4 & \nbeats & \lgbm \\
        \mfive & 30490 & 1913 & 28 & 12 & \lgbm & \lgbm\\
        \bottomrule
    \end{tabular}
    \label{tab:datasets}
\end{table}

\begin{table*}[t]
\centering
% \small
\caption{Test Hierarchical RMSSE ($R_H$) (mean$\pm$std) for different forecast/HPO methods with their reconciliation algorithms in four datasets. The best score is in \textbf{bold}, and the second best is \underline{underlined}. \permbu~failed in \tourL~data. ``Opt. recon." is the abbreviation for optimal reconciliation (\mint~ and \erm).}
\begin{tabular}{c|c|c|c|cccc}
\toprule
Tag & Sub-tag & \makecell{Forecast/HPO\\method} & \makecell{Reconciliation\\algorithm} & \tour & \tourL & \wiki &\traffic \\
\toprule
Gold & HPO on test & \makecell{Student specific\\to dataset} & \naivebu & 0.4668$\pm$0.0117 & 0.4907$\pm$0.0018 & 0.3199$\pm$0.0076 & 0.3736$\pm$0.0103 \\ \midrule
\multirow{11}{*}{ \rotatebox[origin=c]{90}{State-of-the-arts (SOTAs)} } & \multirow{7}{*}{\makecell{Statistical\\with opt. recon.} } & \arima & \naivebu & 0.5434$\pm$0.0000 & 0.5462$\pm$0.0000 & 0.7533$\pm$0.0000 & 0.5353$\pm$0.0000\\
& & \ets & \naivebu & 0.5264$\pm$0.0000 & 0.5204$\pm$0.0000 & 0.7180$\pm$0.0000 & 0.4954$\pm$0.0000\\
% & \arima & \mintshr & 0.5481 & \underline{0.4960}& 0.4282 & 0.4716\\
% & \arima & \mintols & 0.5648 & 0.5386 & 0.5804 & 0.4556\\
& & \arima & \mint & 0.5481$\pm$0.0000 & 0.4960$\pm$0.0000& 0.4282$\pm$0.0000 & 0.4556$\pm$0.0000\\
% & \ets & \mintshr & 0.5134 & 0.5007 & 0.5216 & 0.4733\\
% & \ets & \mintols & 0.5021 & 0.5155 & 0.4455 & 0.4683\\
& & \ets & \mint & 0.5021$\pm$0.0000 & 0.5007$\pm$0.0000 & 0.4455$\pm$0.0000 & 0.4683$\pm$0.0000\\
& & \arima & \erm & 2.8064$\pm$0.0000 & 1.8756$\pm$0.0000 & 0.3940$\pm$0.0000 & 0.9248$\pm$0.0000\\
& & \ets & \erm & 10.2069$\pm$0.0000 &1.9253$\pm$0.0000& 0.4229$\pm$0.0000 & 1.4080$\pm$0.0000\\
% & \thetamodel & \mint & 0.5250$\pm$0.0000 & 0.4903$\pm$0.0000 & 0.4398$\pm$0.0000 & 0.9316$\pm$0.0000 \\ 
% & \thetamodel & \erm & 0.9050$\pm$0.0000 & 1.5465$\pm$0.0000 & \underline{0.3340}$\pm$\underline{0.0000} & 0.8230$\pm$0.0000 \\ 
& & \permbu & \mint & \underline{0.5011$\pm$0.0140} & -- & 0.4244$\pm$0.0436 & 0.4704$\pm$0.0132\\ 
\cmidrule{2-8}
% & \deepar & \mint & 0.5220$\pm$0.0783 & 0.5387$\pm$0.0443 & 0.3373$\pm$0.0325 & 0.8083$\pm$0.2320\\
% & \deepar & \erm & 1.1858$\pm$0.1065 & 1.2219$\pm$0.0873 & 0.3371$\pm$0.0035 & 0.6458$\pm$0.0864\\
% & \nbeats & \mint & 0.5275$\pm$0.0163 & 0.8234$\pm$0.0438 & \textcolor{red}{0.3166$\pm$0.0175} & 0.5382$\pm$0.0035\\
% & \nbeats & \erm & 0.9194$\pm$0.0568 & 1.4614$\pm$0.1232 & 0.3311$\pm$0.0017 & 0.7512$\pm$0.3695\\
% & \deepar & N/A & 0.6236$\pm$0.0419 & 0.5749$\pm$0.0140 & 0.4896$\pm$0.0678 & 0.4352$\pm$0.0229\\
& \multirow{2}{*}{\makecell{End-to-end\\DNN}}  & \deepvarplus & Inherent & 0.6757$\pm$0.0602 & 0.6264$\pm$0.0349 & 0.7527$\pm$0.1476 & 0.4693$\pm$0.0629\\ 
& & \hieretoe & Inherent & 0.5713$\pm$0.0411 & 0.6201$\pm$0.0257& 0.5054$\pm$0.0905 & 0.3910$\pm$0.0217\\ 

\cmidrule{2-8}
& \multirow{2}{*}{\makecell{Teacher model\\with opt. recon.}} & \multirow{2}{*}{\makecell{Teacher specific\\ to dataset}} & \mint & 0.5220$\pm$0.0783 & \textbf{0.4903$\pm$0.0000} & 0.3373$\pm$0.0325 & 0.5382$\pm$0.0035 \\ 
& & & \erm & 1.1858$\pm$0.1065 & 1.5465$\pm$0.0000 & 0.3371$\pm$0.0035 & 0.7512$\pm$0.3695 \\
% & \multirow{4}{*}{\makecell{Teacher with \tcv\\and opt. recon.}} & \multirow{4}{*}{\makecell{Teacher model\\specific to\\dataset}} & \mint & \makecell{\deepar-\mint:\\\textit{0.5220$\pm$0.0783}} & \makecell{\thetamodel-\mint:\\\textit{0.4903$\pm$0.0000}} & \makecell{\deepar-\mint:\\0.3373$\pm$0.0325} & \makecell{\nbeats-\mint:\\0.5382$\pm$0.0035} \\ \cmidrule{4-8}
% & & & \erm & \makecell{\deepar-\erm:\\1.1858$\pm$0.1065} & \makecell{\thetamodel-\erm:\\1.5465$\pm$0.0000} & \makecell{\deepar-\erm:\\\textit{0.3371$\pm$0.0035}} & \makecell{\nbeats-\erm:\\0.7512$\pm$0.3695} \\

\midrule
% ----------------------------------------
\multirow{4}{*}{\rotatebox[origin=c]{90}{Baselines}} & \multirow{4}{*}{\makecell{Student model\\ with \tcv}}  & \tcv-\lowest & \naivebu & 0.8996$\pm$0.2112 & 0.5101$\pm$0.0009 & 0.3904$\pm$0.0609 & 0.4014$\pm$0.0073\\
& & \tcv-\lowest-\po & \naivebu & 0.8313$\pm$0.1558 &  0.5128$\pm$0.0011 & 0.3904$\pm$0.0609 & 0.4077$\pm$0.0146\\
& & \tcv-\hier & \naivebu & 0.6966$\pm$0.1519 & 0.4915$\pm$0.0020 & 0.4439$\pm$0.0504 & 0.4128$\pm$0.0386\\
& & \tcv-\hier-\po & \naivebu & 0.6770$\pm$0.1789 & 0.4997$\pm$0.0013 & 0.4439$\pm$0.0504 & 0.4920$\pm$0.0398\\

\midrule
% & \texttt{Teacher} & \td & 0.6178 && 0.2885 \\  
% & \texttt{Teacher} & \mo~\texttt{(Best)} & 0.5541 && \textbf{0.2277} \\ \midrule
%-----------------------------------------
\multirow{4}{*}{\rotatebox[origin=c]{90}{Ours}} & \multirow{4}{*}{\makecell{\hpro\\variants}} & \hpro-\avg & \naivebu & 0.5310$\pm$0.0223 & \underline{0.4907$\pm$0.0018} & \underline{0.3242$\pm$0.0085} & 0.3827$\pm$0.0093\\
& & \hpro-\avg-\po & \naivebu & \textbf{0.4673$\pm$0.0094} & 0.4935$\pm$0.0005 & \underline{0.3242$\pm$0.0085} & \textbf{0.3766$\pm$0.0034}\\
&  & \hpro-\toplevel & \naivebu & 0.5138$\pm$0.0179 & 0.4953$\pm$0.0055 & \textbf{0.3230$\pm$0.0072} & 0.3869$\pm$0.0070\\
& & \hpro-\toplevel-\po & \naivebu & 0.5158$\pm$0.0178 & 0.4924$\pm$0.0007 & \textbf{0.3230$\pm$0.0072} & \underline{0.3812$\pm$0.0193}\\ \midrule
\multicolumn{4}{c|}{\textbf{Relative improvement w.r.t. best SOTA}} & \textbf{+6.75\%} & \textbf{-0.08\%} & \textbf{+4.18\%} & \textbf{+3.68\%} \\
\multicolumn{4}{c|}{\textbf{Relative improvement w.r.t. best baseline}} & \textbf{+30.97\%} & \textbf{+0.16\%}  & \textbf{+17.16\%} & \textbf{+6.18\%} \\
\bottomrule
\end{tabular}
\label{tab:results}
\end{table*}

We present extensive empirical evaluation on five publicly available datasets: \tour~\cite{tourismdata}, \tourL~\cite{wickramasuriya2019optimal}, \wiki~\cite{wikidumps}, \traffic~\cite{Dua:2019}, and \mfive~\cite{makridakis2022m5}.
The datasets are prepared according to~\cite{rangapuram2021end}.
A summary is given in Table~\ref{tab:datasets}. %, with more details in Appendix~C. 
% \tourL~and \traffic~are abbreviated as \texttt{Tour-L} and \texttt{Traff}, respectively for space constraints.

\subsubsection{Forecasting models} 
Table~\ref{tab:datasets} shows the models employed for different datasets. 
See Section~\ref{sec:Introduction} for references to the models.
For the student, \hpro~works with any ML model that can be regularized by HPO. 
We validate the strength of \hpro~by employing two models as students: \deepar~and \lgbm. 
For \tourL~data, the student is an ensemble of \thetamodel~\cite{assimakopoulos2000theta} and \lgbm.
% , and since \thetamodel~does not generally need any HPO, we tune only \lgbm's hyperparameters in this scenario.
For teacher, \hpro~works with both classical models that do not require HPO and complex ML models needing HPO.
We employ \thetamodel, \deepar, \lgbm, and \nbeats~as teachers in different datasets.
% In this work, we work with a single student and a single teacher for every dataset, but multiple teacher models can also be employed.
We choose both the student and teacher models by assessing their performance on the validation set for every dataset. This approach helps us establish robust baselines, as described in Section~\ref{subsec:Detailed results on benchmark datasets}.
Note that an independent \thetamodel~model is built for each time series, while for all other models, a single model is trained on all time series collected from the suitable levels.
% More details are provided in the appendix. \textcolor{orange}{Small correction - For teachers, we do not use all levels. M5: top 5, tourism-lg: top5, etc. Can we have one small section on teacher selection where we can define H-PRO-TOP and H-PRO-AVG (i.e. top k levels)}  

\subsubsection{Teacher configuration} We explore two teacher configurations.
\begin{enumerate*}[(1)]
    \item In \hpro-\toplevel, proxies from the top-most level is only utilized by the student, \ie $w(1)=1$, and $w(l)=0, \forall 2 \le l \le L-1$.
    \item In \hpro-\avg, proxy forecasts from more than one higher level are utilized by the student. Hence, $w(l)=1/L_T, \forall 1 \le l \le L_T \le L-1$.
\end{enumerate*} 
We set $L_T = L-1$ for \tour, \wiki, and \traffic~data, and $L_T=5$ for \tourL~and \mfive~data.
The reason for choosing lower $L_T$ for \tourL~and \mfive~is that they have relatively deep hierarchies, and training a teacher with data from levels very deep in the tree conflicts with our hypothesis of training the teacher with less sparse data.
% of finding better coherence in the data used for training teacher models.

\subsubsection{Performance metric} 
We adopt the scale-agnostic Root Mean Squared Scaled Error~(RMSSE) as our base metric (used in \mfive~competition).
For a single time series $x^{l,j}$, it is defined as: $r^{l,j} = \sqrt{e/e_{\mathrm{naive}}}$, where $e=\frac{1}{H} \sum_{t=T+1}^{T+H} ( \xlj -\xhatlj )^2$, and $e_{\mathrm{naive}}=\frac{1}{T-1} \sum_{t=2}^{T} ( \xlj - x^{l,j}_{t-1} )^2$.
% \begin{equation}
%     \mathrm{RMSSE}^{l,j}= \sqrt{\frac{ \frac{1}{H} \sum_{t=T+1}^{T+H} \left( \xlj -\xhatlj \right)^2 }{ \frac{1}{T-1} \sum_{t=2}^{T} \left( \xlj - x^{l,j}_{t-1} \right)^2 }}.
% \end{equation}
For multiple series, generally a weighted average is considered. RMSSE at a certain level $l$ is given by, $r^l = \sum_{j}\alpha_j \times r^{l,j}$.
In our experiments, $\alpha_j=1/N_l$ for all datasets except for \mfive.
A special weighting scheme is used in \mfive~as described in \cite{makridakis2022m5}.
As introduced in Section~\ref{sec:Background}, we adopt mean aggregation across levels, and employ \textbf{Hierarchical RMSSE}, $R_H= \frac{1}{L} \sum_{l=1}^{L}r^l$ as our primary metric. A lower value of $R_H$ is preferred. 
Note that \hpro~and other methods implemented here always produce coherent forecasts across hierarchies via reconciliation.

% Hence, we do not check for coherency.}
% \textcolor{blue}{Since all the mentioned approaches generate forecast at all levels via reconciliation,as opposed to seperate forecasts per level, h }

% We also evaluated with other metrics like MAPE, sMAPE, and MASE~\cite{oreshkin2019n}, but due to space constraint, we only report RMSSE in the paper, and one other metric is reported in the appendix. We see similar relative patterns in the performance with all the metrics.

\subsection{HPO framework}\label{subsec:HPO framework}
We adopt Random search~\cite{bergstra2012random} and Hyperopt~\cite{bergstra2013making} as search algorithms, and use RayTune~\cite{liaw2018tune} to perform end-to-end training and HPO in a distributed Kubernetes cluster.
% More details about HPO trials and grids for every model are given in Appendix~C.
We employ two model selection frameworks:
\begin{enumerate*}[(a)]
    \item The \emph{Standard} approach selects the best HPO trial based on the aggregated \tcv/\hpro~objective across the entire forecast horizon.
    \item The \emph{Per-offset~(\po)} method selects the best trial for each offset time-point in the horizon based on \tcv/\hpro~objective at that time, and then concatenates the individual predictions to generate the forecast for the entire horizon.
\end{enumerate*}  
The second method does more granular selection from a pool of HPO trials.
% assigns more importance on the validation/proxy data because of the per-offset selection.

\subsection{Detailed results on benchmark datasets}\label{subsec:Detailed results on benchmark datasets}

\begin{table*}
\centering
\caption{Level-wise mean RMSSE for the best \hpro, best baseline, and best SOTA methods. The best score is \textbf{bolded}, second best is \underline{underlined}. 
% Full table with standard deviation is in Appendix~C. 
``$-$'' denotes unavailability of levels in a dataset.}
\begin{tabular}{c|c|ccccccccc}
\toprule
Data & \makecell{Forecaster} & L1 & L2 & L3 & L4 & L5 & L6 & L7 & L8 \\
\toprule
% -------------------- tourism --------------------
\multirow{3}{*}{\texttt{Tourism}} & Ours (\hpro-\avg-\po) & \textbf{0.3383} & \textbf{0.4251} & \textbf{0.5336} & \textbf{0.5723} &  $-$ &  $-$ &  $-$ &  $-$ & \\
& Best baseline (\tcv-\hier-\po) & 0.6473 & 0.7137 & 0.6836 & 0.6635 &  $-$ &  $-$ &  $-$ &  $-$ & \\
& Best SOTA (\permbu-\mint) & \underline{0.3843} & \underline{0.4766} & \underline{0.5539} & \underline{0.5895} &  $-$ &  $-$ &  $-$ &  $-$ & \\
\midrule
% -------------------- tourismLG --------------------
\multirow{3}{*}{\makecell{\texttt{Tourism}-\texttt{L}}} & Ours (\hpro-\avg) & \underline{0.1812} & \underline{0.3861} & 0.4737 & \underline{0.5749} & \textbf{0.4409} & \textbf{0.5575} & 0.6405 & \textbf{0.6704} & \\
& Best baseline (\tcv-\hier) & 0.1838 & 0.3869 & \underline{0.4732} & 0.5759 & \underline{0.4433} & \underline{0.5579} & \underline{0.6403} & \underline{0.6705} & \\
% & Best SOTA (\arima-\mint) & \textbf{0.1486} & 0.4079 & 0.4986 & 0.5812 & 0.4482 & 0.5757 & \textbf{0.6345} & 0.6732 & \\
& Best SOTA (Teacher-\mint) &  \textbf{0.1534} & \textbf{0.3777} & \textbf{0.4653} & \textbf{0.5692} & 0.4823 & 0.5630	& \textbf{0.6334} & 0.6784 \\
\midrule
% -------------------- wiki --------------------
\multirow{3}{*}{\wiki} & Ours (\hpro-\toplevel) & \underline{0.1954} & \textbf{0.3007} & \underline{0.3256} & \textbf{0.4420} & \underline{0.3515} &  $-$ &  $-$ &  $-$ & \\
& Best baseline (\tcv-\lowest-\po) & 0.3947 & 0.4027 & 0.3502 & 0.4558 & \textbf{0.3485} &  $-$ &  $-$ &  $-$ & \\
% & Best SOTA (\arima-\erm) & \textbf{0.1615} & \underline{0.3132} & \textbf{0.3165} & 0.5827 & 0.5960 &  $-$ &  $-$ &  $-$ & \\
& Best SOTA (Teacher-\erm) & \textbf{0.1514}	& \underline{0.3107} & \textbf{0.2906} & \underline{0.4538} & 0.4790 &  $-$ &  $-$ &  $-$ & \\
\midrule
% -------------------- traffic --------------------
\multirow{3}{*}{\texttt{Traffic}} & Ours (\hpro-\avg-\po) & \textbf{0.1764} & \textbf{0.2103} & \underline{0.2865} & 0.8332 &  $-$ &  $-$ &  $-$ &  $-$ & \\
& Best baseline (\tcv-\lowest) & \underline{0.2324} & 0.2627 & 0.3233 & \textbf{0.7870} &  $-$ &  $-$ &  $-$ &  $-$ & \\
& Best SOTA (\hieretoe) & 0.2329 & \underline{0.2423} & \textbf{0.2726} & \underline{0.8163} &  $-$ &  $-$ &  $-$ &  $-$ & \\
\bottomrule
\end{tabular}
\label{tab:levelwise-result}
\end{table*}

Table~\ref{tab:results} shows \textit{test} Hierarchical RMSSE for \hpro, state-of-the-art, and baseline methods. 
% We show mean and standard deviation (``std'') of the metric over multiple (3/4) HPO runs with different seeds (some classical forecasters have std=0 because of deterministic behavior). 
We show the mean and standard deviation (``std'') of the metric over three experiments ran with three different random seeds (some classical forecasters have std=0 because of deterministic behavior). 
A single HPO run generally contains hundreds of trials. 
% (see Appendix~C).

\subsubsection{Gold} In Table~\ref{tab:results}, the row with tag ``Gold'' shows the result of base (student) forecasters with HPO performed on the \textit{test} set. It gives lower bounds on the errors achievable by the students. Hence, our methods (with tags ``Baselines'' and ``Ours'') should \textit{not} be able to outperform the Gold $R_H$ scores.

\subsubsection{State-of-the-arts (SOTAs)} 
We have three types of SOTAs (marked with three different sub-tags in Table~\ref{tab:results}: (1) Statistical forecasters with optimal reconciliation algorithms, (2) End-to-end deep learning based models that have inherent reconciliaton, and (3) Dataset-specific teacher models with optimal reconciliations.
% Following \cite{rangapuram2021end}, we adopt three reconciliation methods for comparison: naive bottom-up~(\naivebu), \mint, and \erm~(see Section~\ref{sec:Related work}). 
% We present results on multiple combinations of base forecasters and reconciliation methods. 

For statistical methods, we choose two classical forecasters: \arima~and exponential smoothing (a.k.a. \ets), and present results in combination with \naivebu, \mint, and \erm. We experimented with two variants of \mint: ~\mintshr~and \mintols, and report the best one in the paper.
We also present the performance of \permbu~\cite{taieb2017coherent} in combination with \mint.
% For DNN-based methods, we compare with \deepar~without any reconciliation~\cite{salinas2019high}, \deepvarplus~with reconciliation as a post-processing~\cite{rangapuram2021end}, and the recent \hieretoe~method~\cite{rangapuram2021end}.

For DNN-based methods, our benchmarks are \deepvarplus~with reconciliation as a post-processing~\cite{rangapuram2021end}, and the recent \hieretoe~ method \cite{rangapuram2021end}.
For probabilistic forecasters (like \hieretoe), we take the mean forecast as point forecast.

For the third sub-category, we choose the teacher model for a particular dataset (see Table~\ref{tab:datasets}), train it on all time series from all levels, and apply two optimal reconciliation algorithms (\mint~and \erm) on it. 

\subsubsection{Baselines} Our baselines are the direct application of student models along with \bu~reconciliation. \tcv-\lowest~and \tcv-\hier~refer to \tcv-based models targeting the lowest-level's RMSSE and hierarchical RMSSE respectively (see Section~\ref{subsec:HPO with temporal cross-validation (TCV)}). 
\tcv-\lowest-\po~and \tcv-\hier-\po~refer to their extensions with per-offset model section (see Section~\ref{subsec:HPO framework}). We employ single and multiple (up to 4) validation windows for the baselines, and report the best results. 

%%% not giving MO, TD now...
%%%We also present results after directly applying \texttt{Teacher}'s forecast with top-down (\td) and middle-out (\mo) reconciliations. We tried \mo~from all higher levels where teacher was trained, and present the best result in Table~\ref{tab:results}. 
%%% \textcolor{orange}{Remove: MO, TD, references to standard reconciliation methods in all places in this section} 

\subsubsection{Observation on Hierarchical RMSSE}
We build four versions of \hpro~as shown in Table~\ref{tab:results} with ``Ours'' tag: \hpro-\avg~and \hpro-\toplevel~as explained in Section~\ref{subsec:Experimental setting}, and their per-offset extensions \hpro-\avg-\po~and \hpro-\toplevel-\po.
The relative improvements with respect to the best SOTA and the best baseline are shown in the last two rows of Table~\ref{tab:results}.
% We can see that \hpro~outperforms all the baselines and the SOTAs in all four datasets. 
We can see that \hpro~ outperforms all \tcv~ baselines in the four datasets.
\hpro~ outperforms all SOTAs in three datasets (\tour, \wiki, and \traffic).
Only for \tourL~ dataset, the teacher model with \mint~ reconciliation is marginally better than \hpro.
\hpro~ is the second best method there.

Note that \hpro~ achieves this performance with off-the-shelf forecasting models, which highlights its strength as an HPO technique.
Moreover, all four datasets have different characteristics, \eg \traffic~and \tour~have strong seasonality while \wiki~does not. Despite that \hpro~is able provide similar or superior performance. This empirically validates our initial hypothesis that the better predictability at the higher levels can help learn accurate forecasters (teachers) at those levels, which can in turn help regularize lowest-level base (student) forecasters. 
Note that we do not use the teacher after \hpro~is completed, and the student model along with the simplest \naivebu~ reconciliation can produce accurate and coherent forecasts across all levels.
It also results in faster reconciliation because \naivebu~ is multiple times faster than \mint~ and \erm~(see Appendix~\ref{app:A}).

Another observation is that the per-offset (\po) model selection can be helpful sometimes for \hpro~as well as the baseline \tcv~methods. Note that for \wiki, the per-offset extension achieves the same result as the normal version because the horizon, $H=1$.
Comparing \hpro-\toplevel~and \hpro-\avg~, we see that their relative performances vary across datasets. Hence, a detailed study will be presented in Section~\ref{subsec:Detailed studies}.
% \textcolor{orange}{H-PRO-TOP, H-PRO-TOP-PO are not defined. Also, it is good to explicitly mention 4 our our approach with the associated tag-string as it is not coming out clearly. Also, some discussion around top vs avg and its challenges in teacher selection in Table.2  and address them in ablasion study.}

\subsubsection{Observation on level-wise RMSSE} 
Table~\ref{tab:levelwise-result} shows the mean RMSSE for the best variant of \hpro, the best baseline, and the best SOTA method in the above four datasets. 
% For \tour~and \tourL~data, \hpro~outperforms the contenders in most of the levels.
% For \wiki~and \traffic~data, occasionally the best score is achieved by one of the best competing methods.
Out of total 21 levels, our method achieves the best performance in 11 levels, and the second best in 7 out of remaining 10 levels.

\begin{table*}[t]
\centering
\caption{Hierarchical RMSSE, $R_H$ (\mfive~official metric) and level-wise weighted RMSSE for \hpro, SOTA, and baseline (``Base.'') in \mfive~dataset for ``department+store'' ensemble student model. The best score is in \textbf{bold}.} %Full table is in Appendix~C.}
\begin{tabular}{c|c|ccccccccccccc}
\toprule
Tag & Method & $\mathbf{R_H}$ & L1 & L2 & L3 & L4 & L5 & L6 & L7 & L8 & L9 & L10 & L11 & L12 \\ \midrule
Gold & HPO on test & 0.512 & 0.186 & 0.294 & 0.387 & 0.237 & 0.328 & 0.370 & 0.455 & 0.465 & 0.561 & 1.001 & 0.954 & 0.903 \\
\midrule
SOTA & M5 winner & \underline{0.520} & 0.199 & 0.310 & 0.400 & 0.277 & 0.366 & 0.390 & 0.474 & 0.480 & 0.573 & \textbf{0.966} & \textbf{0.929} & \textbf{0.884} \\
\midrule
Base. & \tcv-\hier & 0.534 & 0.230 & 0.327 & 0.410 & 0.280 & 0.363 & 0.403 & 0.483 & 0.489 & 0.580 & 0.999 & 0.951 & 0.899 \\
\midrule
\multirow{4}{*}{Ours} & \hpro-\toplevel & \textbf{0.512} & \textbf{0.186} & \textbf{0.294} & \textbf{0.386} & \textbf{0.237} & \textbf{0.329} & \textbf{0.370} & \textbf{0.456} & \textbf{0.464} & \textbf{0.561} & 1.003 & 0.955 & 0.903 \\
& \hpro-\avg & 0.534 & 0.231 & 0.327 & 0.409 & 0.280 & 0.363 & 0.402 & 0.483 & 0.488 & 0.580 & 1.000 & 0.951 & 0.899 \\
& \hpro-\toplevel-\po & 0.521 & 0.189 & 0.305 & 0.398 & 0.247 & 0.339 & 0.383 & 0.468 & 0.477 & 0.572 & 1.009 & 0.961 & 0.909 \\
& \hpro-\avg-\po & 0.534 & 0.227 & 0.325 & 0.408 & 0.277 & 0.362 & 0.401 & 0.483 & 0.486 & 0.580 & 1.001 & 0.953 & 0.901 \\
\bottomrule
\end{tabular}
\label{tab:m5-result}
\end{table*}

\subsection{Result on large-scale retail forecasting}\label{subsec:Result on large-scale retail forecasting}
Here we validate our method in a large-scale retail forecasting dataset (${\sim}43$K time series, 5.4 years of daily data) from the \mfive~accuracy competition. 
Note that a set of 24 classical benchmark forecasting methods were significantly outperformed by the winners of the competition (see appendix of \cite{makridakis2022m5}), hence, we only compare the performance of \hpro~ with that of the M5 winner method. 
The winning methods in \mfive~demonstrated superior performance by the models built after clustering the data based on certain aggregated level ids.
Hence, we perform ``department''-wise and ``store''-wise clustering, and train one independent \hpro~ model for each cluster. We then ensemble these two forecasts to obtain our final result.  We present the Hierarchical and level-wise RMSSE for the ``department+store'' ensemble model in Table~\ref{tab:m5-result}. 
% Please refer to Appendix~C for detailed individual cluster-wise results. 
We can see that \hpro-\toplevel~outperforms the M5 winning method~\cite{makridakis2022m5} by approximately $2\%$ in the Hierarchical RMSSE ($R_H$), the primary metric used in the competition. It is also close to the Gold number. \hpro-\toplevel~shows superior performance in level-wise RMSSE outperforming the winning method in the top 9 levels. A slight degradation in performance is observed at lower levels, possibly because the effect of the proxy is not being transmitted from the top-most to the lowest level due to the complicated and deep \mfive~hierarchy. Although, we should note that, in the \mfive~competition, the methods that achieved superior performance in the lower levels could not get the same in higher levels, and that was also reflected in their degraded Hierarchical RMSSE scores~\cite{makridakis2022m5}.

\subsection{Discussion and ablation studies}\label{subsec:Detailed studies}

\subsubsection{Teacher performance}
In the above experiments, we select the teacher models through temporal cross-validation with one or (if length permits) multiple validation windows. 
% Teacher model with the best validation RMSSE is chosen.
Table~\ref{tab:result-teacher} shows the level-wise test RMSSE of the selected teachers in different datasets.
In \tour, \traffic, and \wiki, since the level-wise RMSSE of the teacher is better than the \tcv~baseline (from Table~\ref{tab:levelwise-result}), \hpro~gets relatively large improvement ($\sim$ 31\%, 17\%, 6\%) from the baselines.
Similarly, the level-wise teacher performances on the higher levels in those datasets are almost always better than the SOTAs (except for one case: \wiki~L2), which leads to performance improvements of $\sim 7\%, 4\%,$ and $4\%$ respectively.
On the other-hand, for \tourL~and \mfive, where the teacher's level-wise performance is not always superior to the \tcv~baseline, we observe marginal degradation or slight improvement from SOTAs ($\sim -0.1\%$ and $2\%$ respectively for the two datasets). 

An important observation from Table~\ref{tab:result-teacher} and \ref{tab:levelwise-result} is that a student can perform better than its teacher in some of the higher levels, even though it is regularized with the teacher's proxy forecasts. This can be attributed to the student's learning ability from the lowest-level data while regularized by the aggregated signals from the teacher.
\begin{table}[t]
    \centering
    \caption{Teacher's test RMSSE at different higher levels. Levels denoted with ``$-$'' were not used by the teacher.}
    \begin{tabular}{c|ccccc}
        \toprule
        Data & L1 & L2 & L3 & L4 & L5 \\ \midrule
        \texttt{Tourism} & 0.3395 & 0.4297 & 0.5239 & $-$ & $-$ \\
        \texttt{Tourism-L} & 0.2044 & 0.4572 & 0.5458 & 0.6260 & 0.5180 \\
        \wiki & 0.0825 & 0.3291 & 0.2710 & 0.4396 & $-$ \\
        \texttt{Traffic} & 0.1029 & 0.1443 & 0.2299 & $-$ & $-$ \\
        \mfive & 0.1832 & 0.6419 & 0.5850 & 0.4958  & 0.5871 \\ \bottomrule
    \end{tabular}
    \label{tab:result-teacher}
\end{table}

% _----------- OLD wihtout baseline- --------------
% \begin{table}[t]
%     \setlength{\tabcolsep}{5pt}
%     \centering
%     \caption{Hierarchical RMSSE of ensemble models (1) to (4).}
%     \begin{tabular}{c|cccc|c}
%         \toprule
%         % & \multicolumn{4}{|c|}{Ensemble id} & \\ \midrule
%         \makecell{Data} & (1) & (2) & (3) & (4) & SOTA\\ \midrule
%         % Ensemble & \hpro-\toplevel~+\hpro-\avg & \hpro-\toplevel-\po~+ \hpro-\avg-\po & 1+2 & 3+4 \\
%         \texttt{Tour} & 0.5058 & 0.4789 & \textbf{0.4831} & 0.4865 & 0.5011\\
%         \texttt{Tour-L} & 0.4914 & \textbf{0.4882} & \textbf{0.4882} & 0.4883 & 0.4960 \\
%         \wiki & \textbf{0.3229} & - & - & 0.3523 & 0.3940\\
%         \texttt{Traff} & 0.3780 & 0.3732 & \textbf{0.3660} & 0.3703 & 0.3910\\
%         \mfive & 0.5202 & 0.5199 & \textbf{0.5186} & 0.5210 & 0.5213\\
%         \bottomrule
%     \end{tabular}
%     \label{tab:result-ensemble}
% \end{table}
% ---------- end old -------------------
\begin{table}[t]
    \setlength{\tabcolsep}{4pt}
    \centering
    \caption{Hierarchical RMSSE of ensemble models (1) to (4).} %Full table with standard deviations is in Appendix C.}
    \begin{tabular}{c|cccc|cc}
        \toprule
        % & \multicolumn{4}{|c|}{Ensemble id} & \\ \midrule
        \makecell{Data} & (1) & (2) & (3) & (4) & \makecell{Best\\SOTA} & \makecell{Best\\Baseline}\\ \midrule
        % Ensemble & \hpro-\toplevel~+\hpro-\avg & \hpro-\toplevel-\po~+ \hpro-\avg-\po & 1+2 & 3+4 \\
        \tour & 0.506 & \textbf{0.479} & 0.483 & 0.486 & 0.501 & 0.677\\
        \tourL & 0.491 & \textbf{0.488} & \textbf{0.488} & \textbf{0.488} & 0.490 & 0.491\\
        \wiki & \textbf{0.323} & \textbf{0.323} & \textbf{0.323} & 0.352 & 0.337 & 0.390\\
        \traffic & 0.378 & 0.373 & \textbf{0.366} & 0.370 & 0.391 & 0.401\\
        \mfive & 0.520 & 0.520 & \textbf{0.519} & 0.521 & 0.520 & 0.534\\
        \bottomrule
    \end{tabular}
    \label{tab:result-ensemble}
\end{table}

\begin{figure*}[t]
     \centering
     \begin{subfigure}[b]{0.43\columnwidth}
         \centering
         \includegraphics[width=\columnwidth, trim={.8cm .8cm .8cm .8cm},clip]{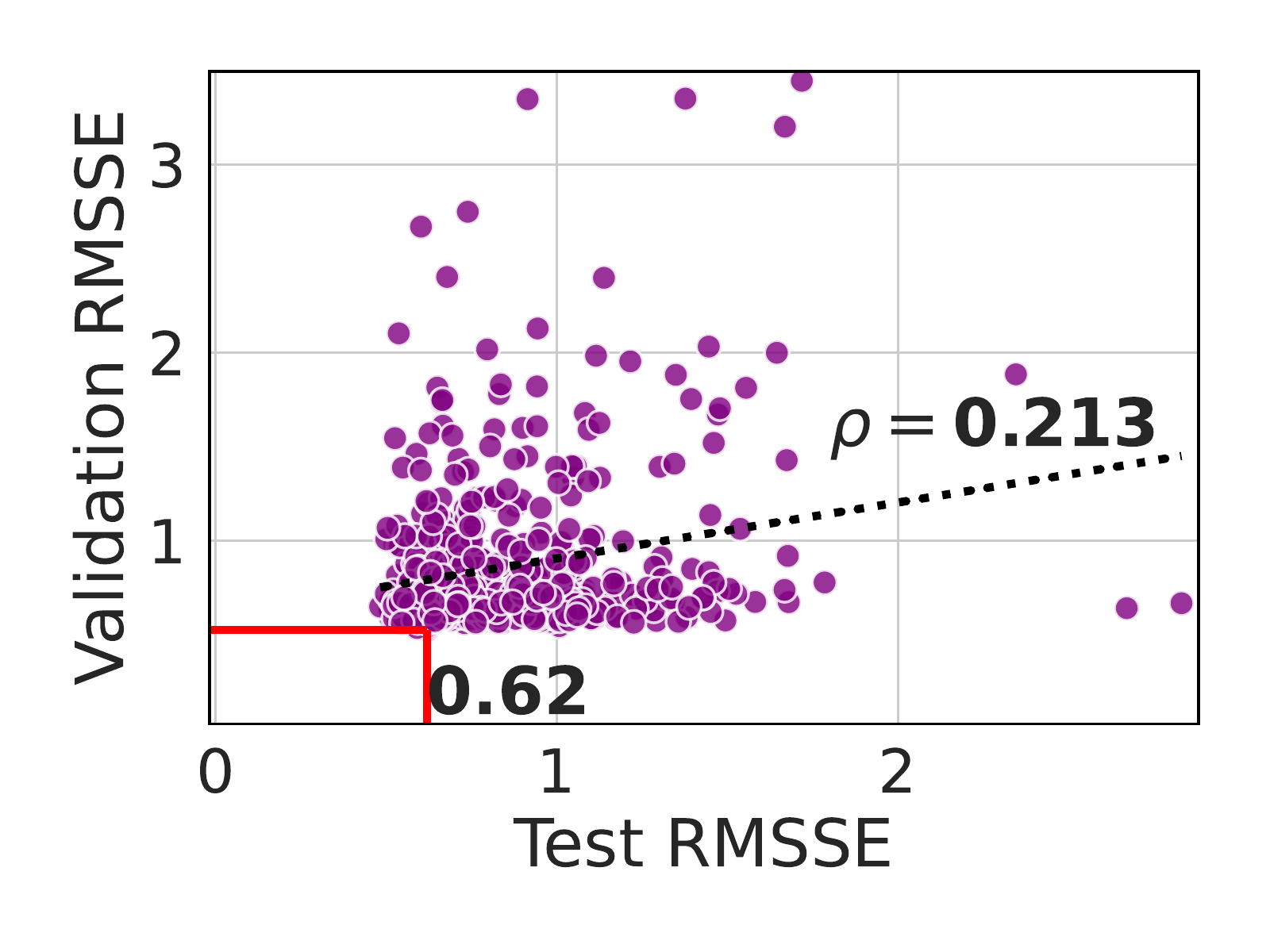}
         \includegraphics[width=\columnwidth, trim={.8cm .8cm .8cm .8cm},clip]{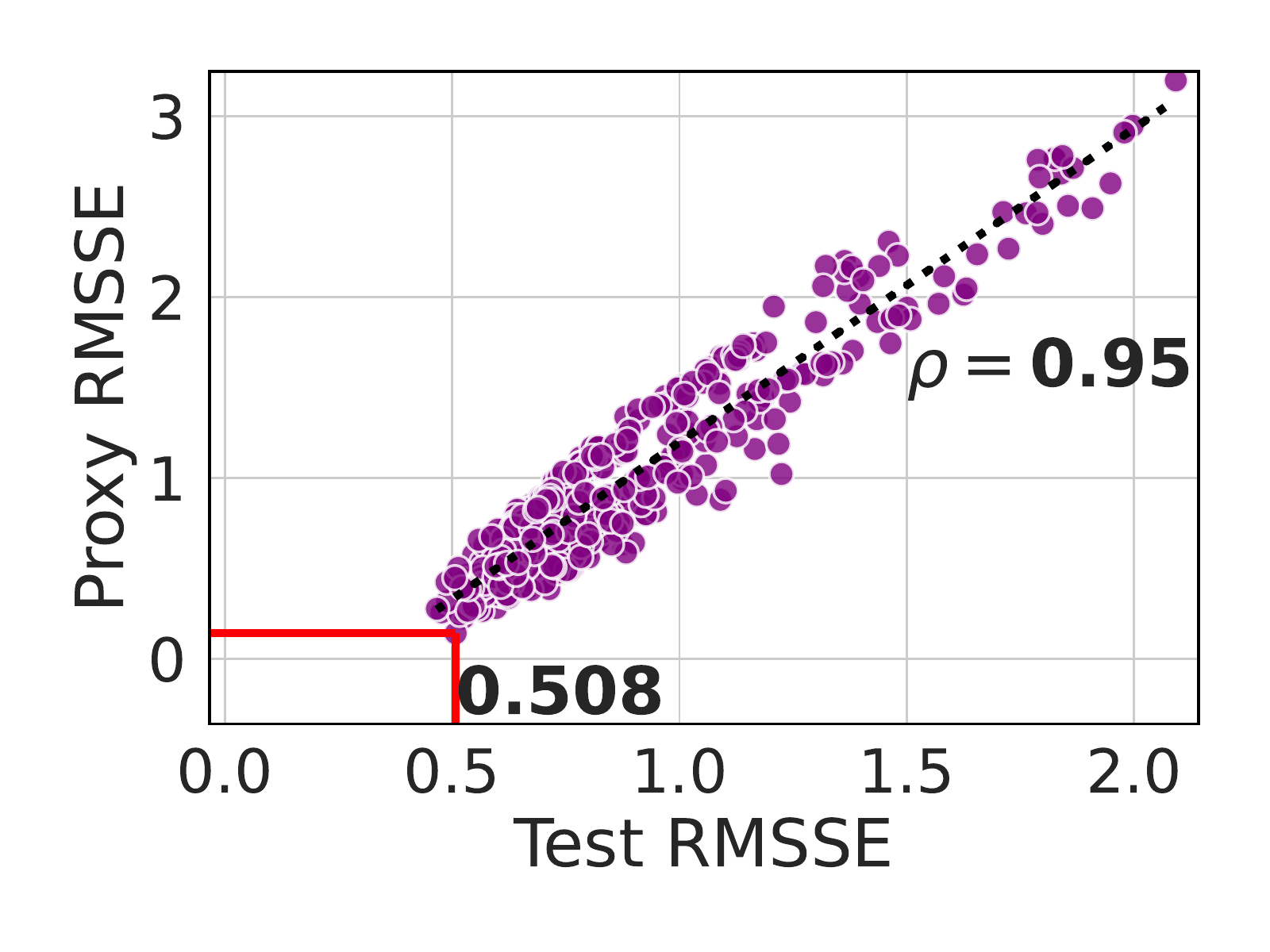}
         \caption{\tour~ data}
         \label{fig:corr-study:tourism}
     \end{subfigure}
     % \hfill
     \begin{subfigure}[b]{0.4\columnwidth}
         \centering
         \includegraphics[width=\columnwidth, trim={1.8cm .8cm .8cm .8cm},clip]{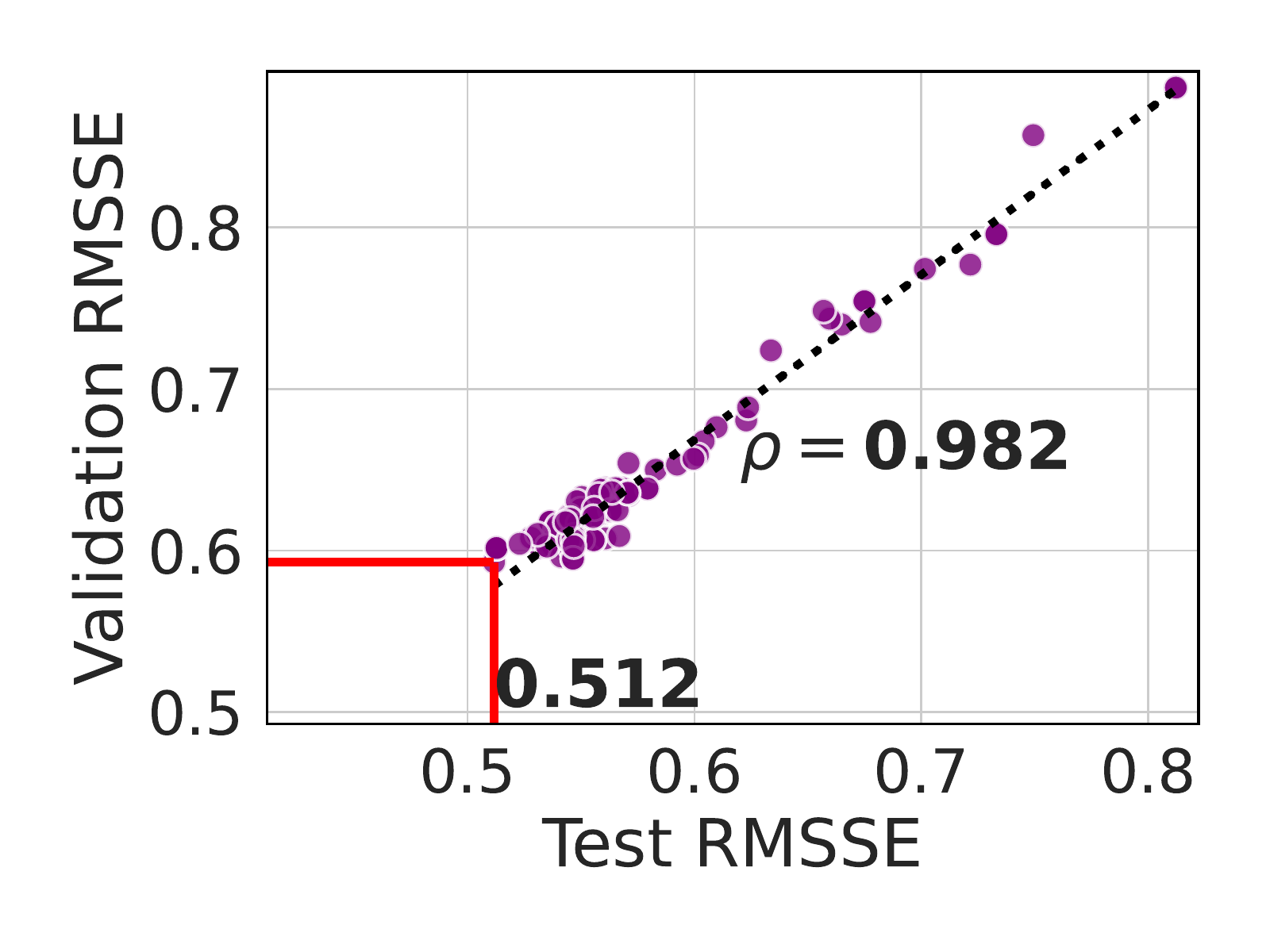}
         \includegraphics[width=\columnwidth, trim={1.8cm .8cm .8cm .8cm},clip]{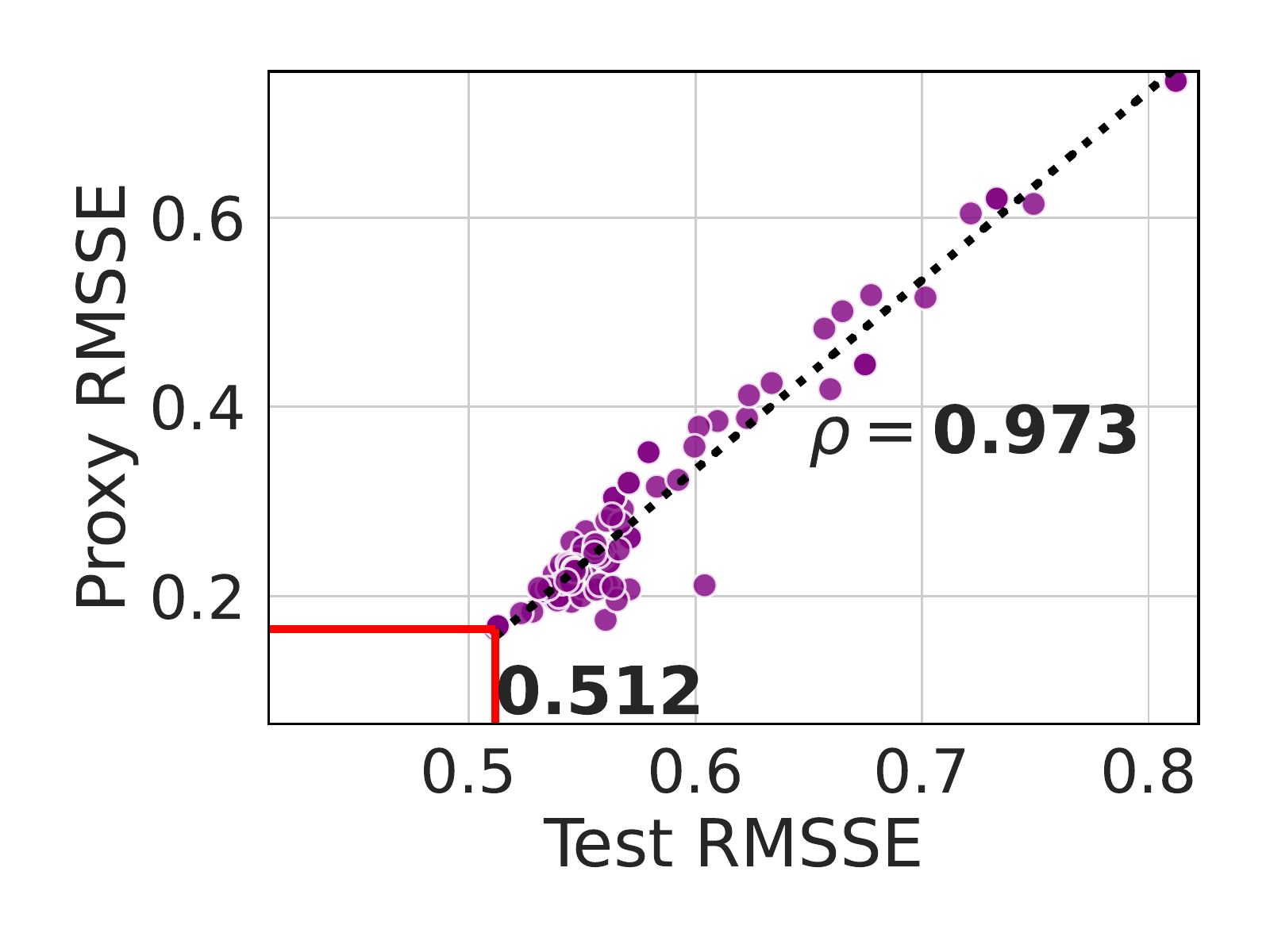}
         \caption{\tourL~ data}
         \label{fig:corr-study:tourism-L}
     \end{subfigure}
     % \hfill
     \begin{subfigure}[b]{0.4\columnwidth}
         \centering
         \includegraphics[width=\columnwidth, trim={1.8cm .8cm .8cm .8cm},clip]{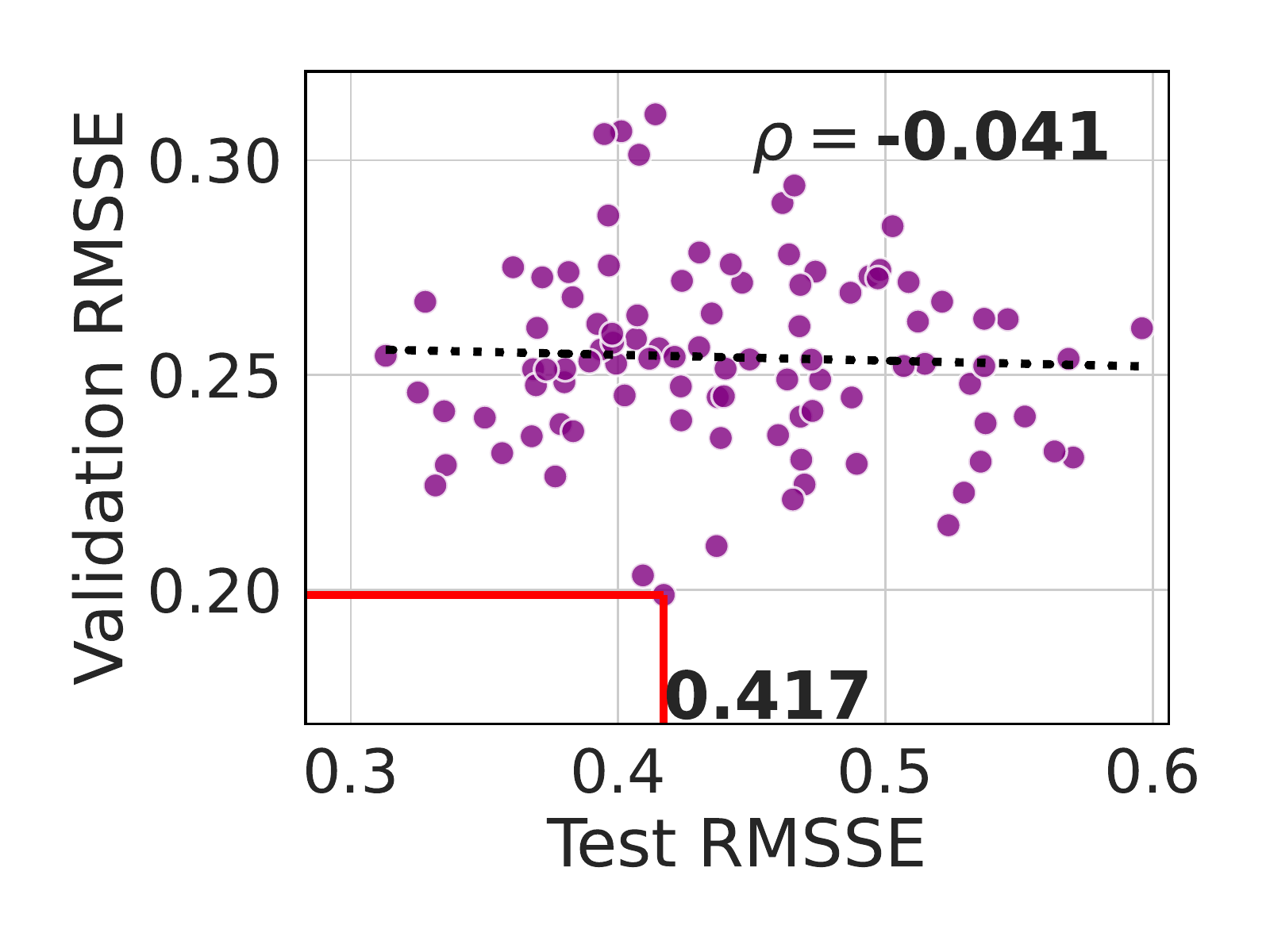}
         \includegraphics[width=\columnwidth, trim={1.8cm .8cm .8cm .8cm},clip]{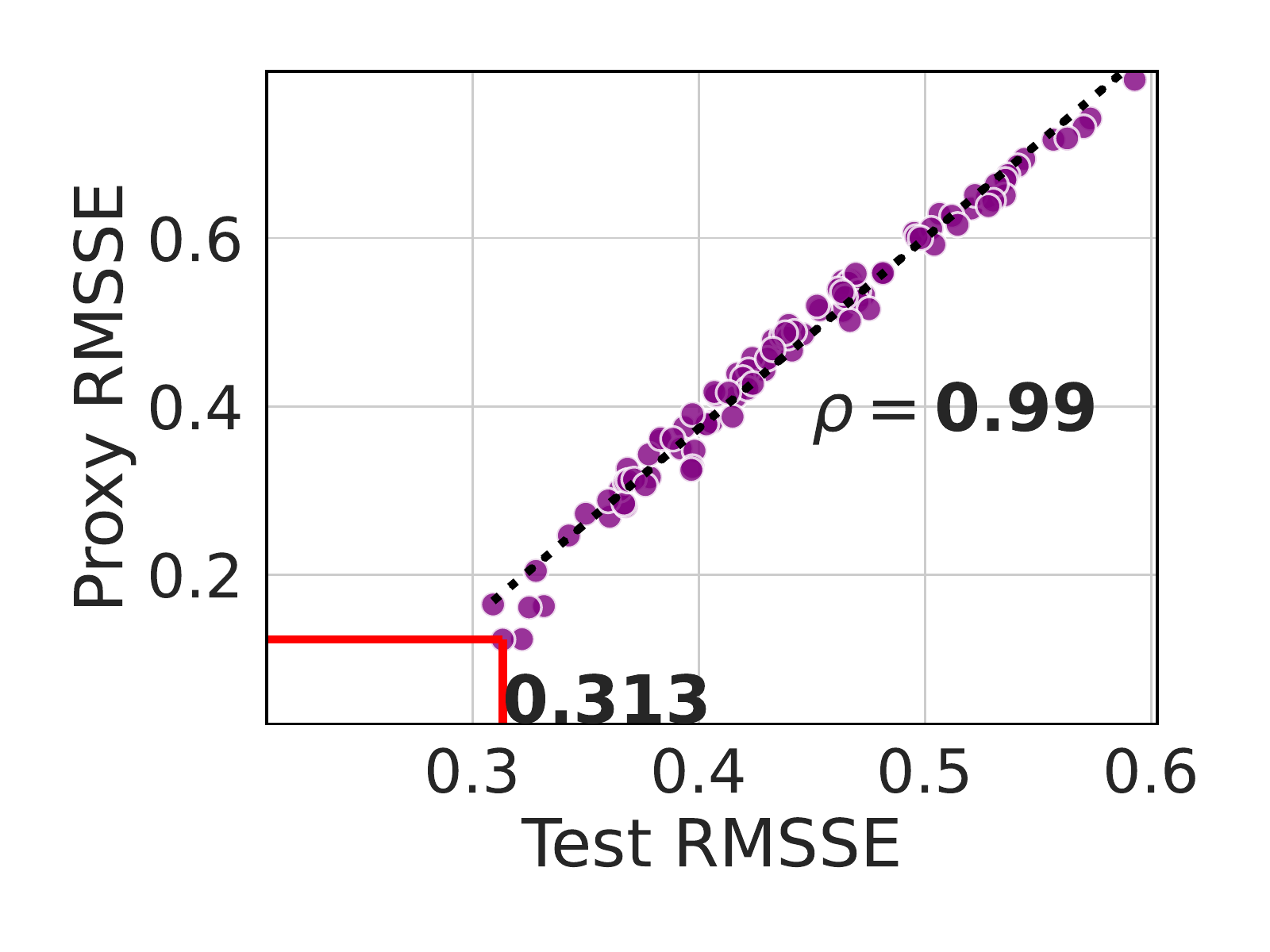}
         \caption{\wiki~ data}
         \label{fig:corr-study:wiki}
     \end{subfigure}
    % \hfill
     \begin{subfigure}[b]{0.4\columnwidth}
         \centering
         \includegraphics[width=\columnwidth, trim={1.8cm .8cm .8cm .8cm},clip]{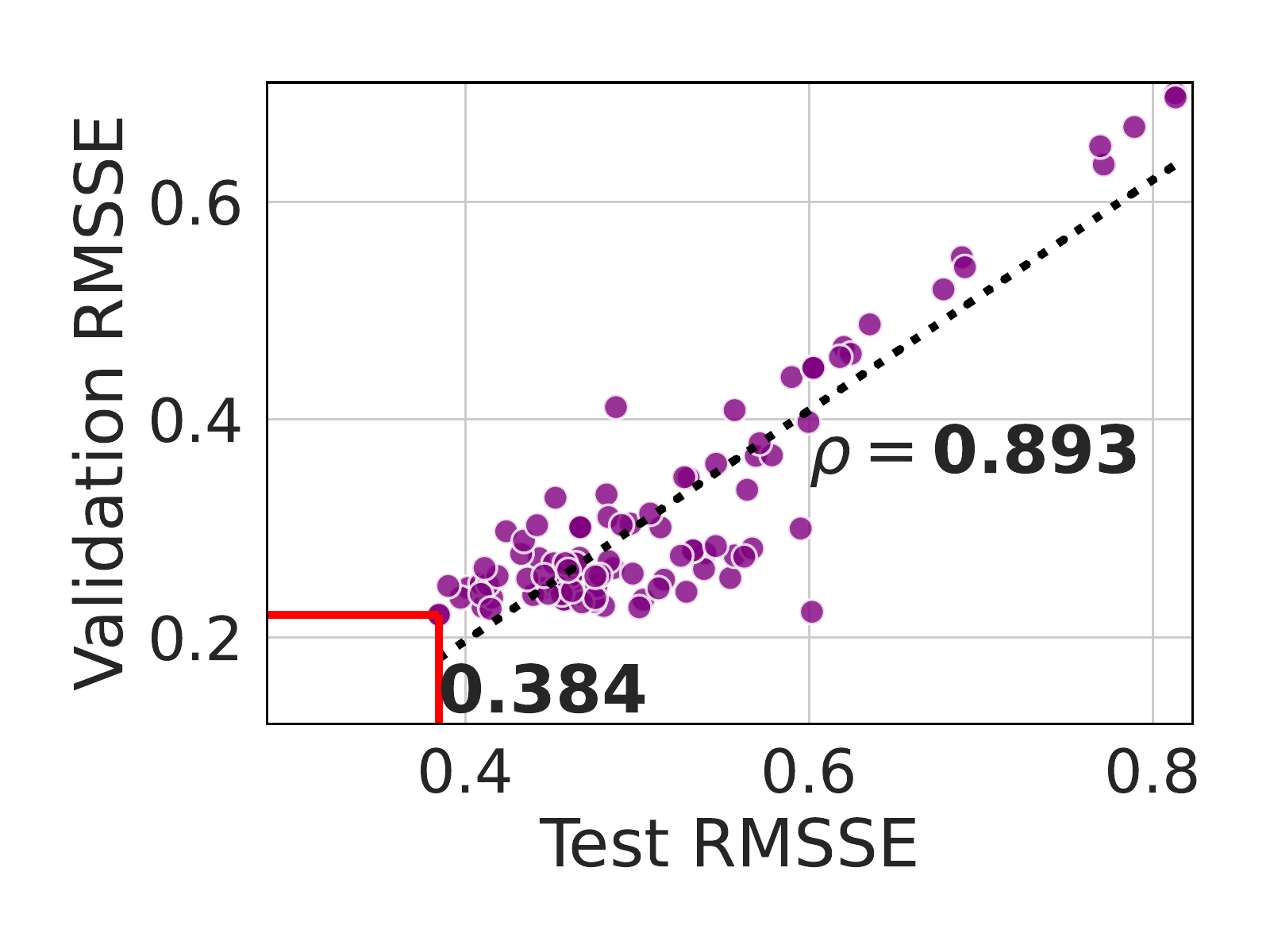}
         \includegraphics[width=\columnwidth, trim={1.8cm .8cm .8cm .8cm},clip]{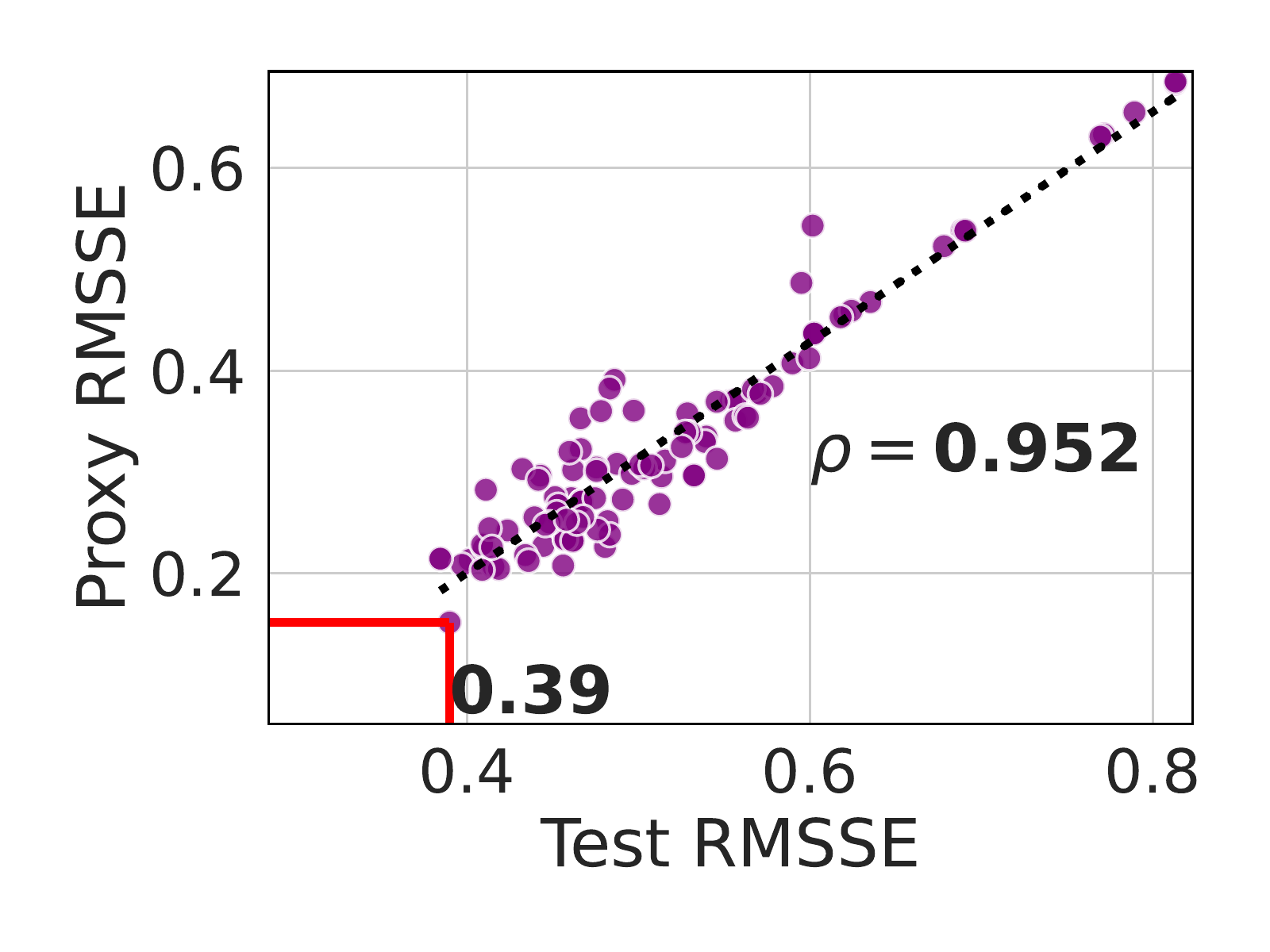}
         \caption{\traffic~ data}
         \label{fig:corr-study:traffic}
     \end{subfigure}
     % \hfill
     \begin{subfigure}[b]{0.4\columnwidth}
         \centering
         \includegraphics[width=\columnwidth, trim={1.8cm .8cm .8cm .8cm},clip]{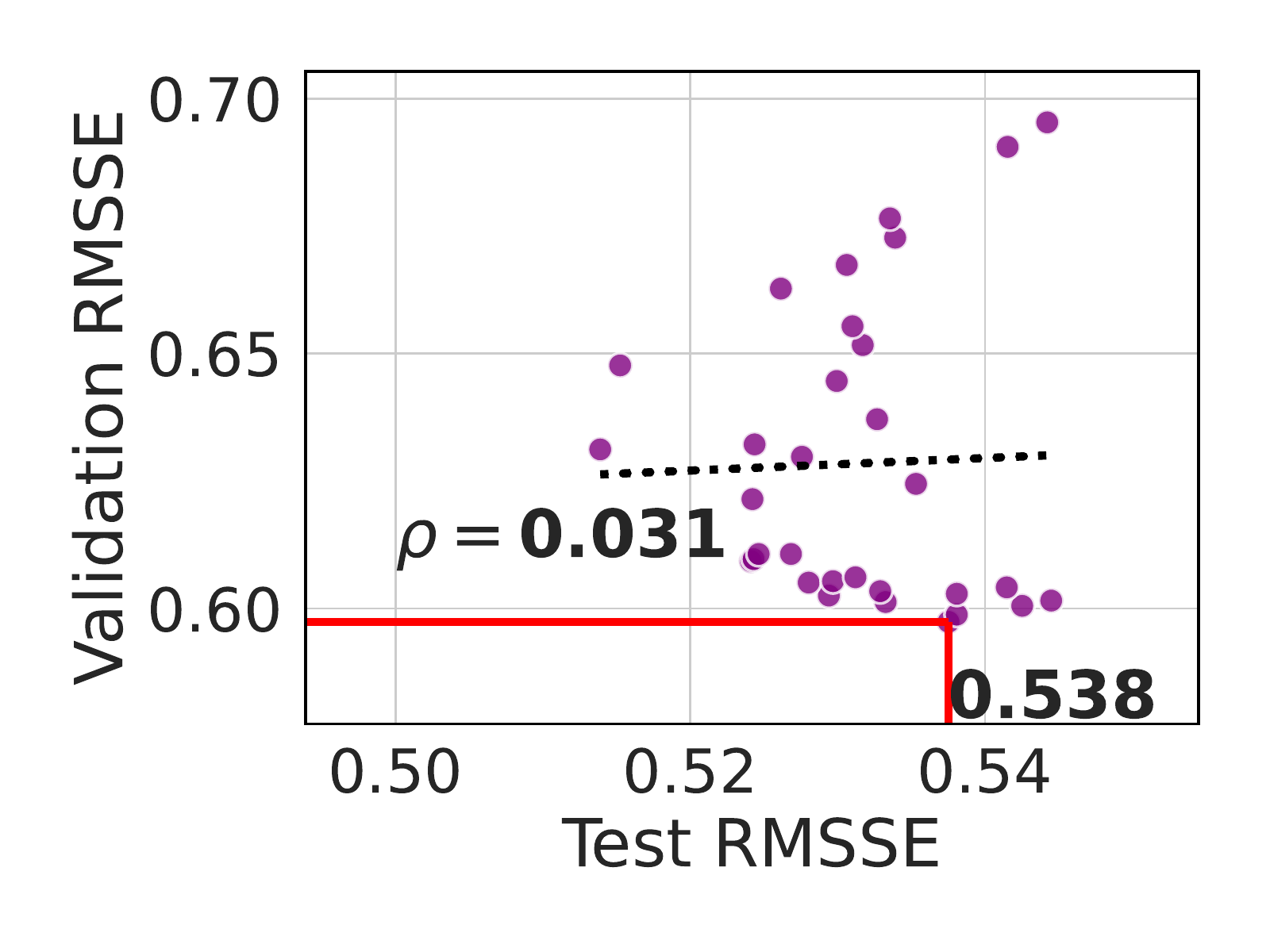}
         \includegraphics[width=\columnwidth, trim={1.8cm .8cm .8cm .65cm},clip]{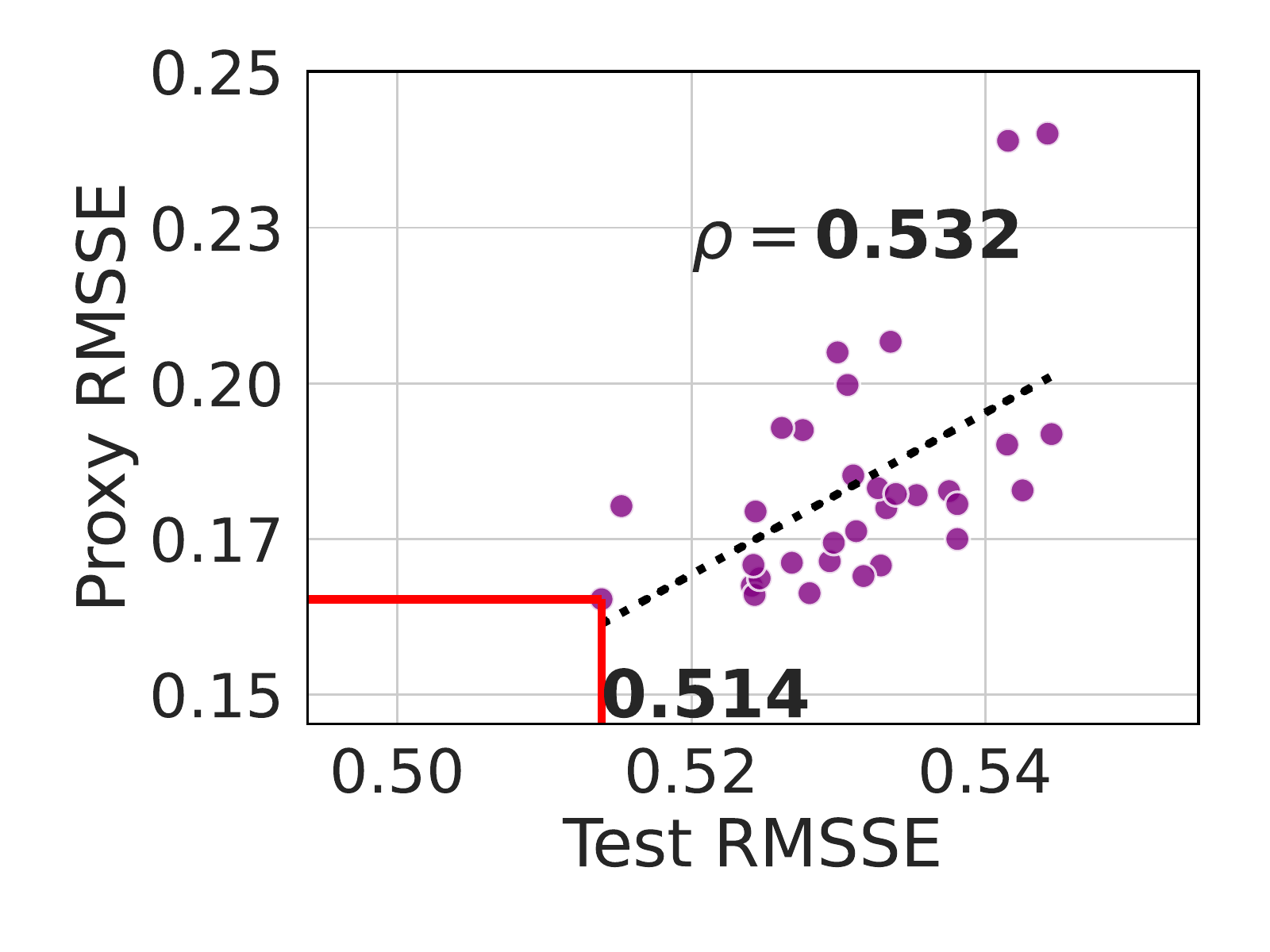}
         \caption{\mfive-Dept. data}
         \label{fig:corr-study:m5-dept}
     \end{subfigure}
     % \hfill
     % \begin{subfigure}[b]{0.34\columnwidth}
     %     \centering
     %     \includegraphics[width=\columnwidth, trim={.8cm .8cm .8cm .8cm},clip]{figures/corr/m5_store/Validation_corr.pdf}
     %     \includegraphics[width=\columnwidth, trim={.8cm .8cm .8cm .8cm},clip]{figures/corr/m5_store/Proxy_corr.pdf}
     %     \caption{\mfive-Store data}
     %     \label{fig:corr-study:m5-store}
     % \end{subfigure}
    \caption{Test errors \textit{vs.} validation errors (upper figure for each subplot). Test errors \textit{vs.} proxy errors (lower figure for each subplot). The solid purple circles denote the HPO trials. Linear trend is shown with the dotted black line, along with Pearson correlation value ($\rho$). The red horizontal line denotes the minimum validation (or, proxy) error, and the corresponding test error is annotated beside the red vertical line. The plots are shown for one random seed. For \mfive-Store, refer to Figure~\ref{fig:intro_motivation_m5}. Best viewed in color.}
    \label{fig:corr-study}
\end{figure*}

% \subsubsection{Effectiveness of HPO}
\subsubsection{Teacher selection and ensemble modeling}
% \textcolor{orange}{Can we explain the difficulty in teacher selection across HPro Top, Avg, PO wrt. above results and then motivate the need for ensembles ?}
As shown in Theorem~\ref{th:2}, an accurate teacher helps the student to produce accurate higher-level forecasts. However, in Section~\ref{subsec:Detailed results on benchmark datasets} and \ref{subsec:Result on large-scale retail forecasting}, we saw that the two variants of \hpro~(\hpro-\avg~and \hpro-\toplevel) built with two configurations of the teacher can achieve the best results interchangeably across datasets. For example, in~\mfive, teacher's test accuracy in L1 is good but poor in other levels. Hence, we observe \hpro-\toplevel~outperforms SOTA while \hpro-\avg~fails, as shown in Table~\ref{tab:m5-result}. In practice, since the teacher's test accuracy is not known, we would need a mechanism to achieve stable performance of~\hpro~\toplevel~and~\hpro-\avg~across datasets.
To address this, we build ensemble models between different variants of \hpro. 
We use the mean of the forecasts from multiple models for ensembles.
As an ablation study, we also build ensemble models between \hpro~and \tcv~baselines. 
We denote these ensembles with the following numbers for concise representation in Table~\ref{tab:result-ensemble}.
\begin{enumerate*}[(1)]
    \item \hpro-\avg~and \hpro-\toplevel, 
    \item \hpro-\avg-\po~and \hpro-\toplevel-\po,
    \item 1 and 2,
    \item 3 and the best \tcv~baseline as obtained in Table~\ref{tab:results}.
\end{enumerate*}
In all datasets, the ensembles perform better than the baselines. 
We can see that the ensemble among all \hpro~variants (id=3) outperforms the best SOTAs and baselines across all five datasets, and hence, can be considered a more stable version of \hpro, which is more robust to %few 
possible sub-optimal teacher performance.
% We can see that the ensemble among all \hpro~variants (id=3) outperforms the best SOTAs and baselines across four out of five datasets, and hence, can be considered as a more stable version of \hpro, which is more robust to few possible sub-optimal teacher performances.
Moreover, ensembles of \hpro~variants and \tcv~improve the performance of the latter, which is beneficial as it shows complementary information was integrated by our approach.

\subsubsection{Correlation study}
Figure~\ref{fig:corr-study} plots the test errors in different HPO trials with respect to validation (or, proxy) errors in those trials for all five datasets used in our experiment.
We present the plots for \tcv-\hier~ and \hpro-\toplevel~ variants.
The linear trend lines are also shown along with Pearson correlation values.
It is evident that \hpro~ obtains much better correlation than \tcv~ for \tour, \wiki, \mfive-Department, and \mfive-Store data (for the last one, refer to Figure~\ref{fig:intro_motivation_m5}).
Hence, HPO trial chosen with the best proxy error often corresponds to better (sometimes the best) test error.
On the other hand, HPO trial selected with the best validation error often results in worse test errors in those scenarios.
For \traffic~ data, in the specific random experiment dictated by the seed, we see that \tcv-\hier~ obtains better test error than \hpro-\toplevel, but the correlation score is higher in the latter.
For \tourL~ data, the scatter plots look similar which is expected from their similar performance (refer to Table~\ref{tab:results}).
Overall, in five out of six scenarios, we see \hpro's proxy error to have better correlation with the actual test error.
This highlights the benefits of our approach in performing HPO with proxy errors, particularly when there is a potential mismatch between the validation and test periods (\eg in \mfive).

\subsubsection{Adaptation to new test window} 
Although, \hpro~is targeted to a particular test-window, we can easily tune it to new test windows by leveraging the saved models from the previous HPO run. \hpro~only requires recomputing the predictions and evaluating the HPO objective for every trial in the past HPO run to select the best model for the new test window. Thus, retraining for all the HPO trials is not mandatory for \hpro, leading to faster adaptation to newer test windows. We should note that, often in forecasting, the immediate past is utilized in training. In that scenario, \hpro~and \tcv~both need to rerun full HPO.

\section{Concluding remarks}
We proposed a hierarchical proxy-guided HPO method for hierarchical time series forecasting to mitigate the perennial problem of data mismatch between validation and test periods in real-life time series.
We provided theoretical justification of the approach along with extensive empirical evidence.
The main benefit of the proposed approach is that it is essentially a model selection or HPO method that can be applied to any off-the-shelf machine/deep learning model for hierarchical time series forecasting. 
% when a notion hierarchy information is available to us, which is often true, either explicitly or implicitly.
We validated \hpro-based HPO with classical machine learning as well as deep learning models in our experiments.
\hpro~ outperformed the conventional temporal cross-validation based HPO approaches in all datasets. 
It also achieves superior results than well-established state-of-the-art methods in four forecasting datasets, and competitive result in one dataset. The performance gain is observed in datasets from diverse domains without requiring any model-specific enhancements.

A future extension can be on formulating a fractional confidence score for the teacher at a certain higher-level node so that the suboptimal teacher forecasts can be given lower priority during the HPO of the student model.
\hpro~ can also be extended to other domains (such as computer vision) when the dataset possesses an inherent hierarchical structure (\eg in hierarchical image recognition).

%%
%% The next two lines define the bibliography style to be used, and
%% the bibliography file.
\bibliographystyle{ACM-Reference-Format}
\balance
\bibliography{aaai23}

%%
%% If your work has an appendix, this is the place to put it.

\appendix

\input{appendix}
\end{document}

%% file: appendix.tex
\appendix

\section*{Appendix}

\section{Advantages of \hpro~over existing reconciliation methods}
\label{app:A}
% Top-down / middle-out reconciliation, although scalable, is typically not as accurate as subsequent more sophisticated reconciliation methods, since it cannot factor in information from the base series forecasts in determining their reconciled forecasts, which has the most granular data about the series and how they might progress.  
State-of-the-art reconciliation methods like \mint~and \erm~ try to factor in forecasts at all levels to derive the final adjusted forecasts, but these have several shortcomings.  First, they are generally not scalable to large number of time series, since they at least require fitting parameter matrices that have size on the order of $N \times N$ where $N$ is the number of base time series.  This fitting essentially requires multiple matrix inversions of matrices of this size (which has complexity more than $O(N^4)$), or for the best performing ERM, solving an even bigger regression problem with $O(TN^2)$ data points (where $T$ is the number of historical time points) and $O(N^2)$ variables.  Additionally they add significant complexity to the forecast process (i.e., getting all hierarchy forecasts on historical data, fitting the reconciliation model, getting forecasts and applying reconciliation model at test time to adjust base forecasts, \etc), and can suffer from overfitting especially with modern ML and DL forecasting approaches that may have close to zero training error, since typically training data forecasts are used to fit the reconciliation model.

Furthermore, both these and the simpler top-down / middle-out reconciliation 
% Furthermore, both these
approaches use fixed linear combinations of different series' forecasts (a single series in the case of top-down and middle-out) to get the adjusted base forecasts, which can be insufficient to accurately predict the base level when the relationship between the levels is more complex (e.g., nonlinear) or changes over time, which is a common case as different local effects can cause proportions relative to aggregates to shift (e.g., consider events like promotion, price change, or advertisement in a retail setting, causing demand and sales for one product to shift to another).

\hpro~on the other hand adjusts the selected base level forecasters directly by leveraging aggregate-level information, hence, can still have time-evolving changes in relative proportions for base level series that factor in all local information. Additionally it avoids having to fit a reconciliation model and apply a complex reconciliation process so it is much more scalable, simpler, and easier to use. While it does require some aggregate level forecasts, these are only needed for the test periods used for model selection.

\section{Proofs}\label{app:B}

\subsection{Proof of Lemma~\ref{th:1}}
This can proved trivially, by employing Definition~\ref{def:Perfect teacher} in \eqref{eq:hpo-obj-proxy-approx2}, and comparing with \eqref{eq:OPT-BU}.

\subsection{Proof of Theorem~\ref{th:2}}
\begin{proof}
Following \eqref{eq:hpo-obj-proxy-approx2},
\begin{equation}
    \mathcal{O} = 
    \frac{1}{L-1} \sum_{l=1}^{L-1} 
    \frac{1}{N_l} \sum_{j=1}^{N_l} 
    \frac{1}{H} \sum_{t=T+1}^{T+H}
    \left( \xtildelj - \xhatlj \right)^2,
\end{equation}
where, we denote $\xhatlj = \mathcal{B} \left( f_\lambda, \left\{ x^{L,i}_{1:T} \right\}_{i=1}^{N_L}, l,j \right)$ for compactness.
Following \eqref{eq:OPT-BU},
\begin{equation}
    \mathcal{O}^\ast = 
    \frac{1}{L-1} \sum_{l=1}^{L-1} \frac{1}{N_l} \sum_{j=1}^{N_l} \frac{1}{H} \sum_{t=T+1}^{T+H}
    \left( \xlj - \xhatlj \right)^2.
\end{equation}

\newcommand{\SSS}[1]{\mathcal{S}\left( #1 \right)}
To make the equations concise, let
\begin{equation}
    \SSS{\cdot} = \frac{1}{L-1} \sum_{l=1}^{L-1} \frac{1}{N_l} \sum_{j=1}^{N_l} \frac{1}{H} \sum_{t=T+1}^{T+H} (\cdot).
\end{equation}

Hence,
\begin{equation}
    \mathcal{O} = 
    \mathcal{S}
    \left( \xtildelj - \xhatlj \right)^2
\end{equation}
\begin{equation}
    \mathcal{O}^\ast = 
    \mathcal{S}
    \left( \xlj - \xhatlj \right)^2.
\end{equation}

Hence,
\begin{align*}
    \biggl| \mathcal{O} \biggl. 
    &- \mathcal{O}^\ast \biggr| \biggr. \\
    &= \left| 
    \SSS  
    {    
    \left( \xtildelj - \xhatlj \right)^2 - 
    \left( \xlj - \xhatlj \right)^2
    } \right| \\
    &= \left| 
    \SSS {       
    \left( \xtildelj - \xlj \right) 
    \left( \xtildelj - 2 \xhatlj + \xlj \right)
    } \right| \\
    &= \left| 
    \SSS {    
    \left( \xtildelj - \xlj \right) 
    \left( 2 \left( \xlj - \xhatlj \right) + \left( \xtildelj - \xlj \right) \right)
    } \right|.
\end{align*}
% Applying triangle inequality ($|a+b|\le |a|+|b|$) three times,
Applying triangle inequality ($|a+b|\le |a|+|b|$),
\begin{align*}
    \biggl| \mathcal{O} \biggl. 
    &- \mathcal{O}^\ast \biggr| \biggr. \\
    &\le 
    \SSS {  
    \left| \left( \xtildelj - \xlj \right)
    \left( 2 \left( \xlj - \xhatlj \right) + \left( \xtildelj - \xlj \right) \right) \right|}\\
    &= 
    \SSS {  
    \left| \xtildelj - \xlj \right|
    \left| 2 \left( \xlj - \xhatlj \right) + \left( \xtildelj - \xlj \right)\right|}.
\end{align*}
Applying triangle inequality again on the inner term,
\begin{align}
    \biggl| \mathcal{O} \biggl. 
    &- \mathcal{O}^\ast \biggr| \biggr. \nonumber\\ 
    &\le 
    \SSS {  
    \left| \xtildelj - \xlj \right|
    \left( \left| 2 \left( \xlj - \xhatlj \right) \right| + \left| \xtildelj - \xlj  \right| \right) } \nonumber\\
    &= \SSS {  
    \left| \xtildelj - \xlj \right|^2 } + 
    2\SSS {  
    \left| \xtildelj - \xlj \right| \left| \xlj - \xhatlj \right| } \nonumber\\
    % &= \mathcal{E} + \frac{2}{L-1} \sum_{l=1}^{L-1}\frac{1}{N_l} \sum_{j=1}^{N_l} \frac{1}{H} \sum_{t=T+1}^{T+H}
    % \epsilon_t^{l,j} \delta_t^{l,j}
    \label{app:eq:O-minus-O-star}
\end{align}
Let $\mathcal{E}$ denote the aggregated mean squared error of teacher's proxy forecasts in all higher levels. Formally, 
\begin{align}
    \mathcal{E} &= \frac{1}{L-1} \sum_{l=1}^{L-1} \frac{1}{N_l} \sum_{j=1}^{N_l} \frac{1}{H} \sum_{t=T+1}^{T+H} \left( \xtildelj - \xlj \right)^2\\
    &= \SSS{ \left| \xtildelj - \xlj \right|^2 }.
    \label{app:eq:MSE-teacher}
\end{align}
Substituting \eqref{app:eq:MSE-teacher} in \eqref{app:eq:O-minus-O-star}, 
\begin{align*}
    \biggl| \mathcal{O} \biggl. 
    &- \mathcal{O}^\ast \biggr| \biggr. \\
    &\le \mathcal{E} + \frac{2}{L-1} \sum_{l=1}^{L-1}\frac{1}{N_l} \sum_{j=1}^{N_l} \frac{1}{H} \sum_{t=T+1}^{T+H}
    \epsilon_t^{l,j} \delta_t^{l,j}.
\end{align*}
\end{proof}

\subsection{Example of reduced variance at higher level}\label{app:Example of reduced variance at higher level}
For the toy hierarchy shown in Figure 2, let $X_1$ and $X_2$ be the time series values at the the leaf nodes, and $Y$ the aggregated sum at the parent. Assume $X_1$, $X_2$ are jointly normal random variables with the following mean and covariance:

\begin{equation}
    \bm{\mu} = \begin{bmatrix} \mu_1 \\ \mu_2 \end{bmatrix}, \\ \quad
    \bm{\Sigma} = \begin{bmatrix}
        \sigma_1^2 & \rho \sigma_1 \sigma_2 \\
        \rho \sigma_1 \sigma_2 & \sigma_2^2
    \end{bmatrix}.
\end{equation}

Then $Y = X_1 + X_2$ is also normally distributed:
\begin{equation}
    Y \sim \mathcal{N} \left( \mu_1+\mu_2, \sigma_1^2+\sigma_2^2+2\rho \sigma_1 \sigma_2 \right)
\end{equation}

If we assume $\sigma_1=\sigma_2=\sigma$, then for $\rho \le -0.5$, 
\begin{equation}
    \sigma_Y^2 \le \sigma^2\,.
\end{equation}